\documentclass[journal]{IEEEtran}
\ifCLASSINFOpdf
\else
\fi
\usepackage{tcolorbox}

\usepackage{cite}
\usepackage{amsmath}
\usepackage{algorithm}
\usepackage{algpseudocode}
\usepackage{graphicx}
\usepackage{ifthen}
\usepackage{caption}
\usepackage{amssymb}
\usepackage{cancel}
\usepackage[utf8]{inputenc}
\usepackage{tabularx}
\usepackage{longtable}
\usepackage{subcaption}
\usepackage{graphicx}
\usepackage[skip=2pt,font=scriptsize]{caption}
\usepackage{enumitem}
\usepackage{hyperref}
\usepackage{booktabs}
\hypersetup{
  colorlinks = true,
  linkcolor = red,
  anchorcolor = red,
  citecolor = green,
  urlcolor = blue,
  pdftitle={Graph-Based Autonomous Vehicle Motion Planning \\Using Game Theory}
}

\begin{document}
\title{Search-Based Autonomous Vehicle Motion Planning \\Using Game Theory}
\author{Pouya~Panahandeh,~Mohammad~Pirani,~Baris Fidan~and~Amir~Khajepour
\thanks{Pouya Panahandeh, Mohammad Pirani, Baris Fidan and Amir Khajepour are with the Mechanical and Mechatronics Eng. Department, University of Waterloo, 200 University Ave W, Waterloo, ON N2L3G1, Canada.(e-mail:\{ppanahan,mpirani,fidan,a.khajepour\}@uwaterloo.ca).}
}
\markboth{Journal of \LaTeX\ Class Files,~Vol.~14, No.~8, April~2024}%
{Shell \MakeLowercase{\textit{et al.}}: Bare Demo of IEEEtran.cls for IEEE Journals}
\maketitle
\begin{abstract}
 In this paper, we propose a search-based interactive motion planning scheme for autonomous vehicles (AVs), using a game-theoretic approach. In contrast to traditional search-based approaches, the newly developed approach considers other road users (e.g. drivers and pedestrians) as intelligent agents rather than static obstacles. This leads to the generation of a more realistic path for the AV. Due to the low computational time, the proposed motion planning scheme is implementable in real-time applications. The performance of the developed motion planning scheme is compared with existing motion planning techniques and validated through experiments using WATonoBus, an electrical all-weather autonomous shuttle bus.
\end{abstract}
\begin{IEEEkeywords}
Motion planning, Game theory, search-based algorithm, Nash equilibrium.
\end{IEEEkeywords}
\IEEEpeerreviewmaketitle
\section{Introduction}
\IEEEPARstart{I}{ntelligent} vehicles have increased their capabilities for highly automated driving under controlled environments i.e.,  driving scenarios that are designed to be predictable, stable, and safe for autonomous vehicles (AVs) to operate in \cite{IEEETransactionsOnMechatronics, IEEETransactionsOnMechatronics2}. Scene information is received using onboard sensors and communication network systems, i.e., infrastructure and other vehicles. Considering the available information, different motion planning and control techniques have been developed for autonomously driving in complex environments. The main goal is focused on executing strategies to improve safety, comfort, and energy optimization. One of the essential conditions for AV safety is ensuring safe interactions with other road users, including human-driven vehicles as well as pedestrians. To this end, this paper proposes an intelligent motion planning algorithm to manage the interaction of the ego vehicle with other road users in different traffic scenarios.

In tackling the interactive motion planning problem, two essential aspects demand careful consideration: the designed waypoints and the velocity profile. These elements play a crucial role in ensuring the effective navigation and safe operation of AV within dynamic environments.
Over the past few decades, numerous AV path planning algorithms have been introduced, however, the majority of these methods do not take into account the interactions between road users. Search-based and sample-based methods are two fundamental branches within the graph-based path planning approaches. 
Search-based path planning algorithms navigate through the search space to find the shortest path between two nodes in a graph by iteratively considering the neighboring nodes, e.g., Dijkstra's algorithm \cite{Fadzli2015}, or they incorporate heuristics to guide the search process efficiently, e.g., A* \cite{Dolgov2010, Erke2020, Yu2020, Lu2022} and D* \cite{Raheem2018} methods. Heuristics are techniques or rules that provide estimated information about the desirability or cost of exploring certain paths, allowing the algorithm to prioritize paths that seem more promising. The approach of sampling-based path planning involves generating numerous potential paths through the space and analyzing them to identify a viable solution. A well-known sampling-based path planning algorithm is called the Rapidly-Exploring Random Tree (RRT) \cite{Veras2019, Li2020, Liao2021, Qi2021, Bazzi2022, Meng2022, Odem2023}. Generally, RRT works by randomly sampling points in the configuration space and constructing a tree of connected points that explores the space in a biased way towards unexplored regions. The algorithm uses heuristics to guide the tree toward the goal region while avoiding obstacles. 
The work of this paper is motivated by the need to consider road user interactions in search-based methods. In the following, we review some approaches on how to address interaction-aware motion planning.

Markov Decision Processes (MDPs) provide a framework to consider the interaction between road users by incorporating probabilistic information about the environment and the road users’ dynamics \cite{Hubmann2018, Hubmann2018-2, Bandyopadhyay2013, Li2019, Sadigh2018, Lauri2023}. In situations where it is not feasible to fully observe the intentions of other drivers, but these intentions can be estimated based on other kinematic features, the Partially Observable Markov Decision Process (POMDP) is utilized. The intentions of surrounding vehicles are considered as hidden variables within the POMDP framework \cite{Intention-aware-online-POMDP}.
Data-driven approaches \cite{Vallon2017, Xiao2022, Zhang2022, Yu2022} for motion planning utilize machine learning techniques and historical data to learn and predict optimal or near-optimal motions for AV. Similarly, Reinforcement learning (RL)-based approaches \cite{Socially-aware-motion-RL, Motion-Planning-Among-Dynamic-RL,Hierarchical-Reinforcement-Learning} require a considerable amount of training data and computation to achieve optimal performance in the interactive environment.

In the literature, there are various works that study the problem of road user interaction, including AVs, as an optimization problem. In other words, the problem is formulated in a way that seeks to find an optimal solution or strategy based on certain objective criteria. The motion planning problem is solved in two stages using mixed-integer linear and nonlinear programming in \cite{Eiras2022}. In \cite{Li2022}, the optimization problem is formed based on artificial potential fields. In \cite{Simon2022}, a solver using the Model Predictive Control (MPC) approach is designed for multiple interacting road users. The interaction is formalized as a root-finding problem.  The quasi-Newton root-finding algorithm and the augmented Lagrangian method are used to solve the problem and handle the constraints, respectively. In \cite{le_cleach2021}, the identification of the road users' objectives is tackled as a prerequisite for solving the optimization problem. The interactive optimization-based approach becomes time-consuming due to its iterative nature, the large size of the search space, and the complexity of handling constraints.

In scenarios involving multiple decision-makers, similar to interactive traffic situations, game theory is a mathematical framework utilized to analyze and model strategic interactions among these decision-makers. It provides a systematic approach for studying scenarios where individual decision outcomes depend not only on their own actions but also on the decisions made by others. In \cite{Dreves2018}, Dreves et al. study traffic scenarios with several AVs, whose dynamics are described by differential equations. To find the solution for the formulated problem they propose the Generalized Nash Equilibrium Problem (GNEP) approach. In \cite{Fridovich2020}, Fridovich-Keil et al. solve a multi-player general-sum differential game exploiting the feedback linearizable structure.  In \cite{Wang2021} and  \cite{Spica2020}, Wang et al. designed a nonlinear receding horizon game-theoretic planner for competitive scenarios between cars and between drones, respectively.  This paper introduces the integration of game theory concepts into a search-based algorithm, resulting in a time-efficient and interactive-aware motion planning algorithm.

The following list outlines the contributions of the paper:
\begin{itemize}
    \item Viewing an interactive traffic situation as a game in which the choices made by individual road users impact the actions of others, and recognizing the efficacy of search-based approaches in illustrating the movement of these road users, we introduce a novel motion planning algorithm by leveraging both the search-based technique and game theory.
    \item By delimiting the search space and incorporating constraints in the game formulation, our new approach reduces the computational time required to find a solution.
    \item Our new approach's adaptability to different vehicle dynamics and road users allows for its extension to a wide range of motion planning problems, including those in robotics and other relevant fields.
\end{itemize}

The paper is structured as follows. In section II, we formulate the problem of motion planning in the context of rational decision-makers interacting. In section III, we introduce the framework of the search-based motion planner utilizing game theory,  the Nash Motion Planner ($N-MP$). In section IV, we investigate an alternative approach within the optimization-based domain for motion planning problem, the Distributed Model Predictive Control Motion Planner ($DMPC-MP$). In section V, we provide a comparative analysis of the two approaches through discussion. Simulation results comparing the performance of the two approaches and the motion planner proposed in \cite{Fridovich2020}, ($ILQ-MP$), for an intersection scenario are presented in section VI. In section VII, we present the experimental results of $N-MP$ applied on WATonoBus, an electrical all-weather autonomous shuttle bus \cite{bhatt2023watonobus}, for a merging scenario. Finally, in section VIII, we summarize the main findings and conclusions of the study, along with highlighting future research directions.

\section{Problem formulation}
When operating in urban traffic, an AV is one among many interacting road users. It needs to plan its future trajectory and speed pattern up to a certain prediction horizon while taking into account the actions of other road users. Consequently, the problem formulation should encompass all road users. Let us denote the 2D position of road user \(i\) at time step \(t\) with \(P_{i(t)} =\) $[ x_{i(t)}, y_{i(t)}]^{T}\in \mathbb{R}^2$ . The orientation and velocity of road user \(i\) are denoted by \(\varphi_{i(t)}\) and \(v_{i(t)}\), respectively. Denoting the state and control input of road user \(i\) as \(X_{i} = [x_{i},y_{i},v_{i},\varphi_{i}]^T\) and \(U_i\), respectively, the joint dynamical system of  \(N\)-road user traffic scenario is derived as
\begin{equation}
X_{(t+1)} = f(X_{(t)},U_{(t)})+X_{(t)}
\label{eqn:dynamic-system}
\end{equation}
where \(X=[X_1^T,..., X_i^T,...,X_N^T]^T\) and \(U=[U_1^T,...,U_i^T,...,U_N^T]^T\). The objective (cost) function of road user \(i\) is denoted by \(J_{i}(X,U_i,U_{-i})\). The objective function $J_i$ depends on the cumulative state \(X\) of all the road users, the control input \(U_i\) of road user $i$, and the other road users' inputs as well, \(U_{-i}=[U_1^T,...,U_{i-1}^T,U_{i+1}^T,...,U_N^T]^T\), due to collision avoidance constraint. We assume that each road user \(i\) has access to the state $X$ but not \(U_{-i}\). Under this assumption, we define the AV trajectory planning problem as an $N$-road user game where each road user $i$ (player $i$), including the AV, aims to minimize its objective while navigating through the traffic environment, i.e., to obtain
\begin{equation}
    \begin{aligned}
        &\underset{X,U_i}{\text{min}}&& J_{i}(X,U_i,U_{-i}).\\
    \end{aligned}
    \label{eqn: Optimization}
\end{equation}
The solution of the $N$-player game formulated above for the ego-AV will be investigated through $N-MP$ and $DMPC-MP$ approach in the following sections.

\section{Nash motion planner}
In this section, the solution to the problem defined in section II is addressed using the $N-MP$ approach. This approach utilizes the search-based method and game theory. In particular, we first define the dynamic equation as explained in \eqref{eqn:dynamic-system}, and then we derive the objective function mentioned in \eqref{eqn: Optimization} and solve it considering the Nash equilibrium. Finally, the speed adjustment related to the AV  is discussed to complete interactive motion planning. The core concept behind this planner is to discretize the search space, which helps accelerate the algorithm's computation time. In the following, these steps will be introduced in detail.
\subsection{Dynamic equation derivation}
In the search-based method, each road user changes its configuration using a displacement vector. Two characteristics related to this vector should be identified: the length and orientation of the vector. The length of displacement vector for road user \(i\) at time step \(t\) is derived as
\begin{equation}
\begin{split}
\Delta P_{i(t)}=K_vv_{i(t)}
\label{length of displacement vector}
\end{split}
\end{equation}
where \(K_v\) is a parameter used for adjusting the relation between the displacement vector length and the longitudinal velocity of the road user \(i\). In an interactive traffic scenario, all road users have similar \(K_v\). Considering accessible movements for all road users, road user \(i\) may select to maintain its primary direction, \(\varphi_{i(t)}\), or change it to some extent. To decrease the computational time of the approach, some limited changes in direction are considered for each road user. Defining $mod(e,m)$ as the integer modulo function that returns  $e$ modulo $m$, i.e., the remainder after the division of $e$ by $m$, the orientation of the displacement vector is calculated as
\begin{equation}
\begin{split}
    \Delta \theta _{i(t)}^{b_i}= 
\begin{cases}
    \varphi_{i(t)} ,& \text{if } \hspace{0.5cm} b_i=1\\
    \varphi_{i(t)} + \frac{b_i}{2} \delta \theta\ ,              & \text{if} \hspace{0.5cm} mod(b_i,2)=0\\
    \varphi_{i(t)}-\frac{b_i-1}{2}\delta \theta\ ,              & \text{else} \hspace{0.5cm}\\
\end{cases}\\
     b_i: 1,2,...,s_i \hspace{2.5cm}
     \end{split}
     \label{orientation of displacement vector}
\end{equation}
where \(\delta \theta\) is a fixed parameter that defines the minimum allowed modification in the direction of road users and \(s_i\) is the maximum number of future nodes for road user \(i\). Defining the displacement vector, we present the road user \(i\) navigation model through the search space as
\begin{equation}
\begin{split}
P_{i(t+1)}^{b_i}=\Delta P_{i(t)}[cos(\Delta \theta _{i(t)}^{b_i}),sin(\Delta \theta _{i(t)}^{b_i})]^T +P_{i(t)}\\
    b_i: 1,2,...,s_i. \hspace{2.5cm}
    \label{dynamic G2MP}
\end{split}
\end{equation}
\subsection{Objective function formulation}
\begin{figure}[h]
    \centering
    \includegraphics[width=5cm]{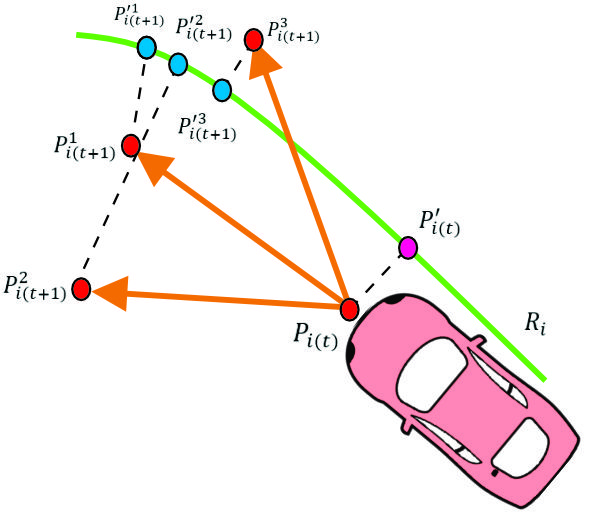}
    \caption{$C_L$ calculation for road user $i$}
    \label{C_L}
\end{figure}

The future configurations of each road user are derived using \eqref{dynamic G2MP}. In the next step, the objective function is defined for each road user. In a general traffic scenario, it is expected that the AV and other vehicles aim to be close to their reference path. \(C_L\) penalizes distancing from the reference path and a shorter distance from the reference path results in a lower cost. Considering \(R_i\) as the reference path of road user \(i\), $\mathcal{D}_{{P_1},{P_2}}$ to be the distance between points $P_1$ and $P_2$, and $proj(P,\mathcal{R})$ to be the orthogonal projection of point $P$ to the path $\mathcal{R}$, the \(C_L\) calculation is summarized in Algorithm \ref{CL} for road user \(i\) who tries to leave its current node, \(P_{i(t)}\), and moves to the next node, \(P_{i(t+1)}\), where it has multiple options to select, \( \boldsymbol{P_{i(t+1)}} = [P_{i(t+1)}^1,P_{i(t+1)}^2,...,P_{i(t+1)}^{s_i-1},P_{i(t+1)}^{s_i}]\). An example is shown in Figure \ref{C_L} for $C_L$ derivation of road user \(i\) who has 3 configurations to select as its future position. The projection of current and future positions of road user $i$ are displayed on the green reference path by pink and blue points. 
\begin{algorithm}
\caption{\(C_L\) calculation}\label{alg:cap}
\begin{algorithmic}[1]
\State\(b_i: 1,2,...,s_i\).
\If{ There is no reference path for the road user $i$:} 
   \State \(C_L(P_{i(t+1)}^{b_i})=0\) \vspace{0.2cm}.
\Else{ There is a reference path, $R_i$, for the road user $i$:} 
    \State Find \(P_{i(t+1)}^{'b_i}=proj(P_{i(t+1)}^{b_i},R_i)\).\vspace{0.2cm}
    \State Find \(P_{i(t)}^{'}=proj(P_{i(t)},R_i)\).\vspace{0.2cm}
    \State $C_L(P_{i(t+1)}^{b_i})=\mathcal{D}_{P_{i(t)},P_{i(t)}^{'}}\mathcal{D}_{P_{i(t+1)}^{b_i},P_{i(t+1)}^{'b_i}}$.
    \EndIf 
\end{algorithmic}
\label{CL}
\end{algorithm}
\graphicspath{ {Images/} }
\captionsetup{justification=centering}

To avoid collision between road users, \(C_D\) is defined to manage the distance between the road users to prevent accidents. \(C_D\) corresponds to the distance between the probable future nodes of road users and is formed in a way that a longer distance results in a lower cost. Considering a general case where road user \(i\) wants to change its position from \(P_{i(t)}\) to \(P_{i(t+1)}\)  in a \(N\)-road user interactive traffic scenario, \(C_D\) is calculated as 
\begin{equation}
\begin{split}  C_D(P_{i(t+1)}^{b_i}|P_{1(t+1)}^{b_1},...,P_{i(t+1)}^{b_i},...,P_{N(t+1)}^{b_N})= \\
    (\sum_{l=1}^{N} \mathcal{D}_{P_{i(t)},P_{l(t)}}\mathcal{D}_{P_{i(t+1)}^{b_i},P_{l(t+1)}^{b_l}})^{-1}. \hspace{0.5cm}\\
    i: 1,2,...,N  \text{  and  }  b_i:1,2,...,s_i. \hspace{0.5cm}
\end{split}
\end{equation} 
As an example, consider a scenario involving two road users in a traffic situation where the first road user has two positions to select at time step $t+1$, $P_{1(t+1)}^1$ and $P_{1(t+1)}^2$, and the second road user has also two positions to pick, $P_{2(t+1)}^1$ and $P_{2(t+1)}^2$. If the first road user selects $P_{1(t+1)}^1$ and on the other side the second road user selects $P_{2(t+1)}^1$, $C_D(P_{1(t+1)}^{1}|P_{1(t+1)}^{1},P_{2(t+1)}^{1})$ objective is calculated as $(\mathcal{D}_{P_{1(t)},P_{2(t)}}\mathcal{D}_{P_{1(t+1)}^1,P_{2(t+1)}^1})^{-1}$.

The other objective that is common in search-based algorithms is the cost to the goal point, \(C_G\). This objective is important in the case that the road user does not have any reference path, instead, it has a goal configuration to reach. To calculate the \(C_G\), the distance between possible choices of each road user and its goal position is measured and objective functions are formed in a way that lower costs correspond to shorter distances. Considering $G_i$ as the goal of road user $i$, the \(C_G\) calculation procedure is defined in Algorithm \ref{Algorithm C_G}.
\begin{algorithm}
\caption{\(C_G\) calculation}\label{alg:cap}
\begin{algorithmic}[1]
\State\(b_i: 1,2,...,s_i\).
\If{ There is no goal destination for the road user $i$:} 
   \State \(C_G(P_{i(t+1)}^{b_i})=0\).
\Else{ There is a goal destination, $G_i$, for the road user $i$:} 
    \State
$C_G(P_{i(t+1)}^{b_i})=\mathcal{D}_{P_{i(t)},G_{i}}\mathcal{D}_{P_{i(t+1)}^{b_i},G_{i}}$.
    \EndIf 
\end{algorithmic}
\label{Algorithm C_G}
\end{algorithm}

To form the objective defined in \eqref{eqn: Optimization}, once all objective functions are computed, we derive the final objectives corresponding to the future nodes of road user \(i\) in a \(N\)-road user interactive traffic scenario by summing up \(C_L\), \(C_D\), and \(C_G\) as
\begin{equation}
\begin{split}  J_{i}(P_{i(t+1)}^{b_i}|P_{1(t+1)}^{b_1},...,P_{i(t+1)}^{b_i},...,P_{N(t+1)}^{b_N})= \\
C_G(P_{i(t+1)}^{b_i})+C_L(P_{i(t+1)}^{b_i})+\hspace{1.5cm}\\
C_D(P_{i(t+1)}^{b_i}|P_{1(t+1)}^{b_1},...,P_{i(t+1)}^{b_i},...,P_{N(t+1)}^{b_N}).\\
i: 1,2,...,N  \text{  and  }  b_i:1,2,...,s_i. \hspace{1cm}
\end{split}
\end{equation}

\subsection{Nash equilibrium identification}
In order to find the optimal strategy for all road users in the traffic scenario, including AV, the Nash equilibrium is investigated in this part. A set of strategies $(P_{1(t+1)}^*,...,P_{i(t+1)}^*,...,P_{N(t+1)}^*)$ for $N$ road users constitutes a Nash equilibrium if no road user has a unilateral incentive to deviate from his or her strategy. In other words, the following inequality must hold for each road user \(i\):
\begin{equation}
\begin{split}
 J_{i}(P_{i(t+1)}^{*}|P_{1(t+1)}^{*},...,P_{i(t+1)}^{*},...,P_{N(t+1)}^{*}) \leq \\J_{i}(P_{i(t+1)}^{b_i}|P_{1(t+1)}^{*},...,P_{i(t+1)}^{b_i},...,P_{N(t+1)}^{*})\\
 b_i:1,2,...,s_i. \hspace{2cm}
\label{Nash}
\end{split} 
\end{equation}

\begin{table}[!h]
\begin{center}
\caption {Bi-matrix game for two players.}
\label{Table:bi-matrix game}
\begin{tabular}{|c||c|c|} 
\hline
 \textbf{Node} & $P_{2(t+1)}^1$ & $P_{2(t+1)}^2$\\
 \hline\hline
 $P_{1(t+1)}^1$ & [$J_{1}^{11},J_{2}^{11}$] & [$J_{1}^{12},J_{2}^{21}$]\\ 
 \hline
$P_{1(t+1)}^2$ & [$J_{1}^{21},J_{2}^{12}$] & [$J_{1}^{22},J_{2}^{22}$] \\  
\hline
\end{tabular}
\end{center}
\end{table}
As an example, consider the case discussed for $C_D$ calculation between two road users in a $2$-road user interactive traffic scenario. Different combinations of their strategies at time step $t+1$ form a bi-matrix game as represented in Table \ref{Table:bi-matrix game}. In this table $J_{1}(P_{1(t+1)}^{1}|P_{1(t+1)}^{1}, P_{2(t+1)}^{1})$ is demonstrated as $J_{1}^{11}$. In each cell, the first entry is the overall objective related to the first road user and the second one is related to the second road user. There are several ways to identify Nash equilibrium, including cell-by-cell inspection, iterated removal of dominated strategies, and best-response analysis. To find all Nash equilibrium using cell-by-cell inspection, each cell of the bi-matrix game is checked to see if it meets the definition of a Nash equilibrium in (\ref{Nash}). The algorithm examines every cell in the matrix and assesses whether any road user can achieve a higher level of benefit by considering the choices made by the other road user. When no road user can improve their objective function, the algorithm marks that cell as an equilibrium. Conversely, if any road user can enhance their outcome, the algorithm marks that cell as not an equilibrium. A similar procedure is followed for scenarios with more than 2 road users. 

By finding the Nash equilibrium in each step, the optimal strategies $(P_{1(t+1)}^*,...,P_{i(t+1)}^*,...,P_{N(t+1)}^*)$ are identified for all road users. It is worth mentioning that in the problem of motion planning, we are only concerned with the actions taken by the AV. There is a possibility that no Nash equilibrium can be found inside the interactive traffic scenario. This issue will be addressed in future works where we follow other types of methods, e.g., mixed strategies and Stackelberg equilibrium, to find the solution to the game. 
\subsection{Ego-AV speed modification}
One of the factors that influence the interactive traffic scenario is the longitudinal velocity of each road user. It is assumed that the speed of all road users other than the AV remains constant for horizon \(H\). AV modifies its velocity in a way to avoid collision with the closest road user in the traffic scenario. Considering $P_{e(t)}=[x_{e(t)},y_{e(t)}]^T$, \(\varphi_{e(t)}\) and \(v_{e(t)}\) as the position, orientation, and longitudinal velocity of the AV on 2D space at time step $t$, respectively, and $P_{n(t)}=[x_{n(t)},y_{n(t)}]^T$, \(\varphi_{n(t)}\) and \(v_{n(t)}\) as the position, orientation, and longitudinal velocity of the closest road user to the AV, respectively. In Algorithm \ref{Algorithm: speed modification}, we present how the AV changes its velocity. The dot product between vectors is represented as $dot(\hspace{0.2cm},\hspace{0.2cm})$ in this algorithm.

\begin{algorithm}
\caption{Speed modification of AV}\label{alg:cap}
\begin{algorithmic}[1]
   \State Calculate the probable collision point, $P_{c(t)}=[x_{c(t)},y_{c(t)}]^T$, of two road users by the following \cite{Jiménez2013}:
   \begin{gather*}
x_{c(t)} = \frac{(y_{n(t)}-y_{e(t)})-(x_{n(t)}tan(\varphi_{n(t)})-x_{e(t)}tan(\varphi_{e(t)}))}{tan(\varphi_{e(t)})-tan(\varphi_{n(t)})}\\
y_{c(t)} = \frac{(x_{n(t)}-x_{e(t)})-(y_{n(t)}cot(\varphi_{n(t)})-y_{e(t)}cot(\varphi_{e(t)}))}{cot(\varphi_{e(t)})-cot(\varphi_{n(t)})}\\
\end{gather*}
\State Compute the time it takes for the non-AV road user to reach the potential collision point:

$t_n=\mathcal{D}_{P_{n(t)},P_{c(t)}}/v_{n(t)}$
\State Calculate the future position of AV traveled by the time \(t_n\): 

$P_{f(t)}=v_{e(t)}t_n[cos(\varphi_{e(t)}),sin(\varphi_{e(t)})]^T+P_{e(t)}$.
\State Calculate the vector connecting $P_{c(t)}$ to $P_{f(t)}$:

$\overrightarrow{W_c}=P_{c(t)}-P_{f(t)}$

\State Calculate the unit vector of the AV orientation:

$\overrightarrow{W_e}= [cos(\varphi_{e(t)}), sin(\varphi_{e(t)})]^T$
\State Modify the AV velocity for the next time step:

\textbf{if} $\mathcal{D}_{P_{c(t)},P_{f(t)}} \prec Z$ and $dot(\overrightarrow{W_e},\overrightarrow{W_c})\geq 0$ \textbf{then}

\vspace{0.1cm}\hspace{1.5cm}\(v_{e(t+1)}=v_{e(t)}-\Delta v\)

\textbf{if} $\mathcal{D}_{P_{c(t)},P_{f(t)}} \prec Z$ and $dot(\overrightarrow{W_e},\overrightarrow{W_c}) \prec 0$ \textbf{then}

\vspace{0.1cm}

\hspace{1.5cm}\(v_{e(t+1)}=v_{e(t)}+\Delta v\)

\vspace{0.1cm}

\textbf{else} 
\hspace{1.0cm}\(v_{e(t+1)}=v_{e(t)}\)      

 \vspace{0.1cm}
 
 \textbf{endif} 
 
\end{algorithmic}
where \(Z\) is the safe space considered around the non-AV road user and \(\Delta v\) is the allowable changes in the AV speed.  
 \label{Algorithm: speed modification}
\end{algorithm}

\begin{figure}[h]
    \centering
    \includegraphics[width=7cm]{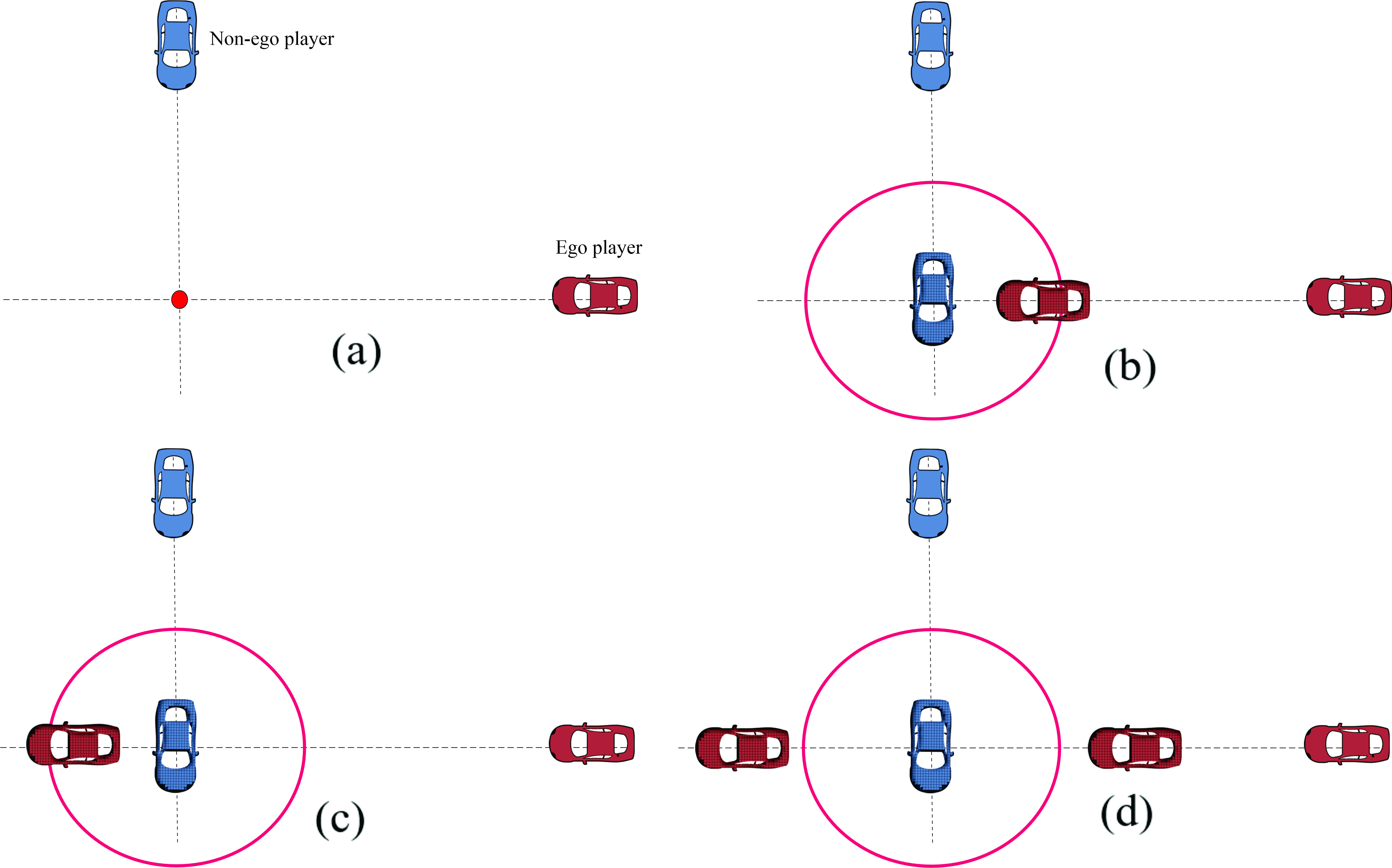}
    \caption{Speed modification for AV}
    \label{Figure: speed modification for AV}
\end{figure}

The rationale behind Algorithm \ref{Algorithm: speed modification} is that the AV always needs to keep itself out of the safe space related to the closest road user to AV. The AV changes its velocity based on its relative distance from the closest road user when this road user reaches the potential collision spot. When the potential location of the accident between road users is identified, the red circle in Figure \ref{Figure: speed modification for AV}(a), three cases are possible. If the non-AV road user arrives at the collision spot sooner than the AV and their relative distance is less than $Z$, the AV needs to decelerate to keep itself far from the other road user, Figure \ref{Figure: speed modification for AV}(b). If the non-AV road user reaches the potential collision spot and the AV has already passed that spot and still their relative distance is less than $Z$, the AV should accelerate, Figure \ref{Figure: speed modification for AV}(c), and if their relative distance is greater than $Z$, there is no need for AV to change its velocity, Figure \ref{Figure: speed modification for AV}(d). 

Changing the velocity of AV influences the magnitude of the displacement vector in equation \eqref{length of displacement vector}. Finding the changes in the velocity of AV in each time step together with identifying its optimal strategy completes the motion planning problem. In the next section, the problem related to motion planning for AV in an interactive traffic scenario will be solved using the $DMPC-MP$ approach. The $DMPC-MP$ approach serves as a baseline for evaluating the performance of $N-MP$ in terms of planned interactive motion, runtime, and theoretical aspects. There will be a comprehensive discussion on $N-MP$ and $DMPC-MP$ in Section V.
\graphicspath{ {Images/} }
\captionsetup{justification=centering}

\section{$DMPC$ motion planner}
This section presents the $DMPC-MP$ approach for addressing the motion planning problem, taking into account interactions between the AV and other road users. This method provides a feedback Nash equilibrium solution, as outlined in \cite{Simon2022}, and will serve as a baseline for evaluating its performance against $N\_MP$. Notably, $DMPC-MP$ operates in a continuous space while searching for the game solution, offering a key point of comparison. In the first step, we need to derive the dynamic equation in \eqref{eqn:dynamic-system}. In $DMPC-MP$ approach, we investigate the case where AV interacts with vehicles in the traffic scenario, however, the same procedure is applicable in the case that AV interacts with pedestrians as well. The nonlinear continuous-time equation that describes a kinematic bicycle model is selected to present the motion of vehicles as
\begin{equation}
\begin{split}
\dot x=vcos(\varphi + \beta) \hspace{1cm}\\
\dot y=vsin(\varphi + \beta) \hspace{1cm}\\
\dot \varphi = \frac{v}{L_r}sin(\beta) \hspace{1cm}\\
\dot v = a \hspace{2cm}\\
\beta = tan^{-1}(\frac{L_r}{L_r+L_f}tan(\delta))
\end{split}
\end{equation}
where \(x\) and \(y\) are the coordinates of the center of the vehicle’s mass in an inertial frame. \(\varphi\) is the inertial heading and \(v\) is the speed of the vehicle. \(L_f\) and \(L_r\) represent the distance from the vehicle's center of mass to the front and rear axles, respectively. \(\beta\) is the angle of the current velocity of the center of mass with respect to the longitudinal axis of the car. \(a\) is the acceleration of the center of mass in the same direction as the velocity. The control inputs are the front steering angles, $\delta$, and the acceleration of the vehicle, \(a\). Considering an interactive traffic scenario that includes the AV and human-driven vehicle, the motion of these vehicles is explained as
\begin{equation}
X_{(t+1)}=f(X_{(t)},U_{e(t)},U_{h(t)})+X_{(t)}
\label{Dynamic: AV and Human driven vehicle}
\end{equation}
where $X=[X_e^T, X_h^T]^T$, \(X_e\) is the state of the AV consisting of \([x_{e},y_{e},v_{e},\varphi_e]\) and \(X_h\) is the state of the human-driven vehicle which includes \([x_{h},y_{h},v_{h},\varphi_h]\), \(U_e\) is the control input of the AV, \(U_e=[\delta_{e},a_e]^T\), and \(U_h\) is the control input of the human-driven vehicle, \(U_h=[\delta_{h},a_h]^T\). The objective considered for both vehicles is that they try to follow the reference path, and simultaneously avoid each other utilizing a repulsive term in their objective function, and also aim to complete their mission with minimum input effort. The objective function of the AV for a prediction horizon $H$ is defined as
\begin{equation}
\begin{split} 
J_e = \frac{1}{2} ((\bold{x_{ed}}-\bold{x_{e}})^{T}\bold{Q_{e1}}(\bold{x_{ed}}-\bold{x_{e}})\\+(\bold{y_{ed}}-\bold{y_{e}})^{T}\bold{Q_{e2}}(\bold{y_{ed}}-\bold{y_{e}})\\
+\bold{a_{e}}^{T}\bold{P_{e1}}\bold{a_{e}}+\bold{\delta_{e}}^{T}\bold{P_{e2}}\bold{\delta_{e}}\\
+{K_e}((\bold{x_e}-\bold{x_h})^{T}(\bold{x_e}-\bold{x_h})\\+(\bold{y_e}-\bold{y_h})^{T}(\bold{y_e}-\bold{y_h})+R_e)^{-1})
\end{split}
\end{equation}

and similarly for human-driven vehicle as
\begin{equation}
\begin{split} 
J_h = \frac{1}{2} ((\bold{x_{hd}}-\bold{x_{h}})^{T}\bold{Q_{h1}}(\bold{x_{hd}}-\bold{x_{h}})\\+(\bold{y_{hd}}-\bold{y_{h}})^{T}\bold{Q_{h2}}(\bold{y_{hd}}-\bold{y_{h}})\\
+\bold{a_{h}}^{T}\bold{P_{h1}}\bold{a_{h}}+\bold{\delta_{h}}^{T}\bold{P_{h2}}\bold{\delta_{h}}\\
+{K_h}((\bold{x_e}-\bold{x_h})^{T}(\bold{x_e}-\bold{x_h})\\+(\bold{y_e}-\bold{y_h})^{T}(\bold{y_e}-\bold{y_h})+R_h)^{-1})
\end{split}
\end{equation}
where $\bold{x_{hd}}=[x_{hd}^1,...,x_{hd}^H]^T$ and $\bold{y_{hd}}=[y_{hd}^1,...,y_{hd}^H]^T$ are the reference path for human-driven vehicle, $\bold{x_{ed}}=[x_{ed}^1,...,x_{ed}^H]^T$ and $\bold{y_{ed}}=[y_{ed}^1,...,y_{ed}^H]^T$ are the reference path for AV, $\bold{x_{h}}=[x_{h}^1,...,x_{h}^H]^T$ and $\bold{y_{h}}=[y_{h}^1,...,y_{h}^H]^T$ are the position of human-driven vehicle,  $\bold{x_{e}}=[x_{e}^1,...,x_{e}^H]^T$ and $\bold{y_{e}}=[y_{e}^1,...,y_{e}^H]^T$ are the position of AV, $\bold{a_{h}}=[a_h^1,...,a_h^H]^T$ and $\bold{\delta_{h}}=[\delta_{h}^1,...,\delta_{h}^H]^T$ are the control inputs from human-driven vehicle, $\bold{a_{e}}=[a_e^1,...,a_e^H]^T$ and $\bold{\delta_{e}}=[\delta_{e}^1,...,\delta_{e}^H]^T$ are the control inputs from AV. $\bold{Q_{e1}}, \bold{Q_{e2}}, \bold{P_{e1}}, \bold{P_{e2}}, \bold{Q_{h1}}, \bold{Q_{h2}}, \bold{P_{h1}}, \bold{P_{h2}}$ are weighting matrices. \(K_e\), \(K_h\), \(R_e\), and \(R_h\) are coefficients that form the repulsive term on objective functions. Obtaining the objectives for both road users, we are able to form the optimization problem in \eqref{eqn: Optimization} as
\begin{equation}
\begin{aligned}
\min_{X,U_e}J_e(X,U_e,U_h)\\
\min_{X,U_h}J_h(X,U_h,U_e)\\  
\end{aligned}
\end{equation}

\begin{figure}[h]
    \centering
    \includegraphics[width=3.5cm]{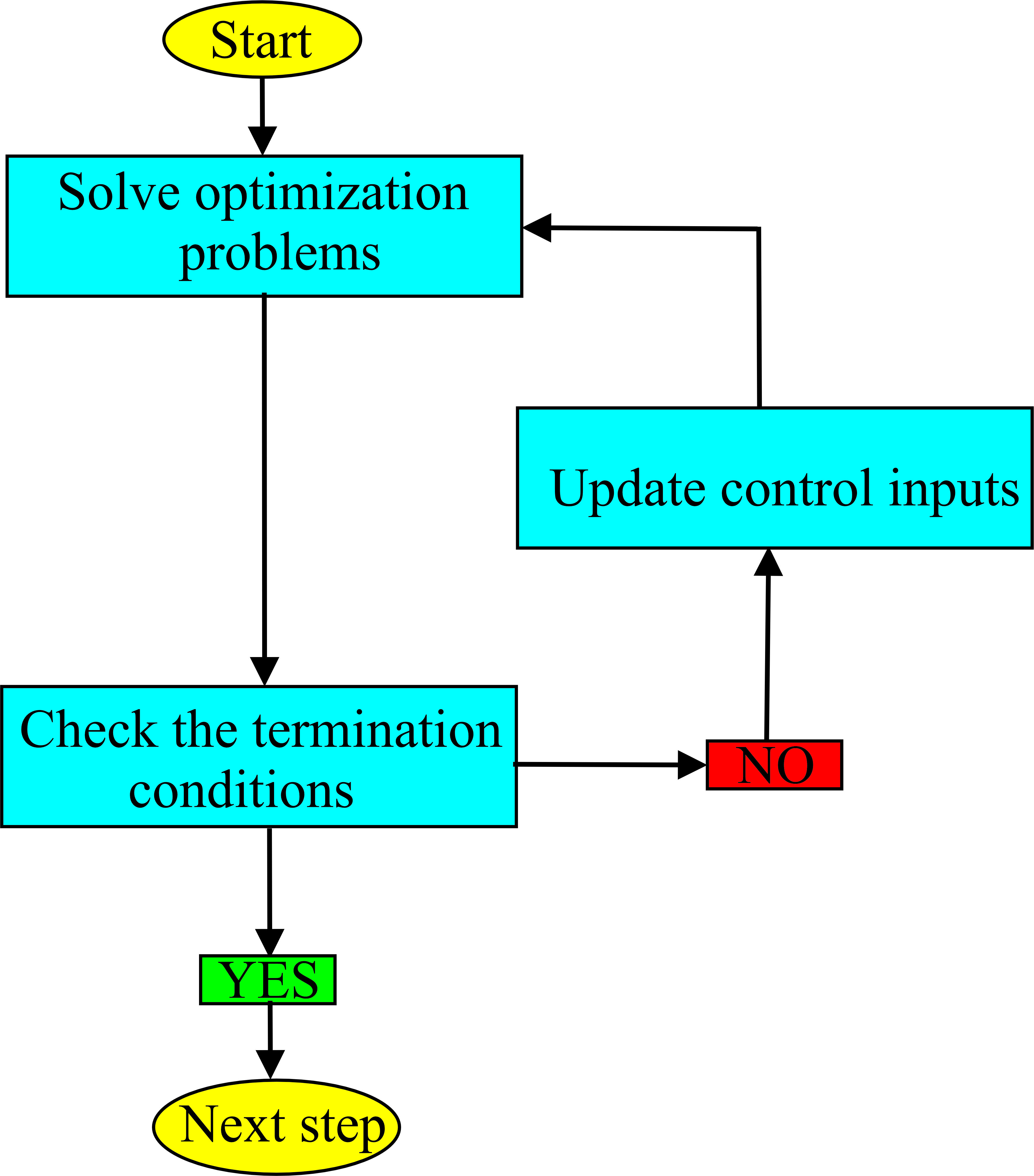}
    \caption{$DMPC-MP$ procedure}
    \label{Figure: DMPC procedure}
\end{figure}

Using the approach proposed in \cite{Christofides2013}, in each iteration, two optimization problems are solved considering the action of the other road user. It is worth mentioning that the action of the human-driven vehicle mainly affects the AV through the repulsive term considered on the AV’s objective function and vice versa. In the next iteration, the termination criterion needs to be checked \cite{Christofides2013}. If the maximum iteration is reached or differences between the control inputs are less than the predefined thresholds, \(\varepsilon_{h}\), \(\tau_{h}\), \(\varepsilon_{e}\), and \(\tau_{e}\), the procedure is over, and outputs are obtained. The convergence threshold among control inputs will be investigated in the following equation where $q$ represents the number of iterations:
\begin{equation}
    \begin{split}
        ||a_h^q-a_h^{q-1}||\leq \varepsilon_{h}\\
        ||a_e^q-a_e^{q-1}||\leq \varepsilon_{e}\\
        ||\delta_h^q-\delta_h^{q-1}||\leq \tau_{h}\\
        ||\delta_e^q-\delta_e^{q-1}||\leq \tau_{e}
    \end{split}
\end{equation}
If the maximum number of iterations is not reached and the differences between control inputs are above the defined thresholds, the procedure needs to be repeated for another iteration. The control inputs for both road users are updated for the next iteration as
\begin{equation}
    U^{q+1}=\alpha U^q + (1- \alpha)U^{q-1}
\end{equation}
where \(\alpha\) is the updating factor. $DMPC-MP$ solves optimization problems iteratively related to the interactive traffic scenario to find the optimized solution for all road users \cite{Christofides2013}. In a $2$-road user traffic scenario, it considers the action from the human-driven vehicle and solves the optimization problem related to AV, similarly, it takes into account inputs from AV and solves the optimization problem related to the human-driven vehicle. If the obtained control inputs for both road users do not differ significantly from the previous iteration, this indicates that the solution obtained is an optimal set of strategies. The overall procedure of the $DMPC-MP$ is represented in Figure \ref{Figure: DMPC procedure}.
\graphicspath{ {Images/} }
\captionsetup{justification=centering}
\section{Discussion}
\begin{figure}[htp]
\centering
\subfloat[Single step]{%
  \includegraphics[clip,width=0.4\columnwidth]{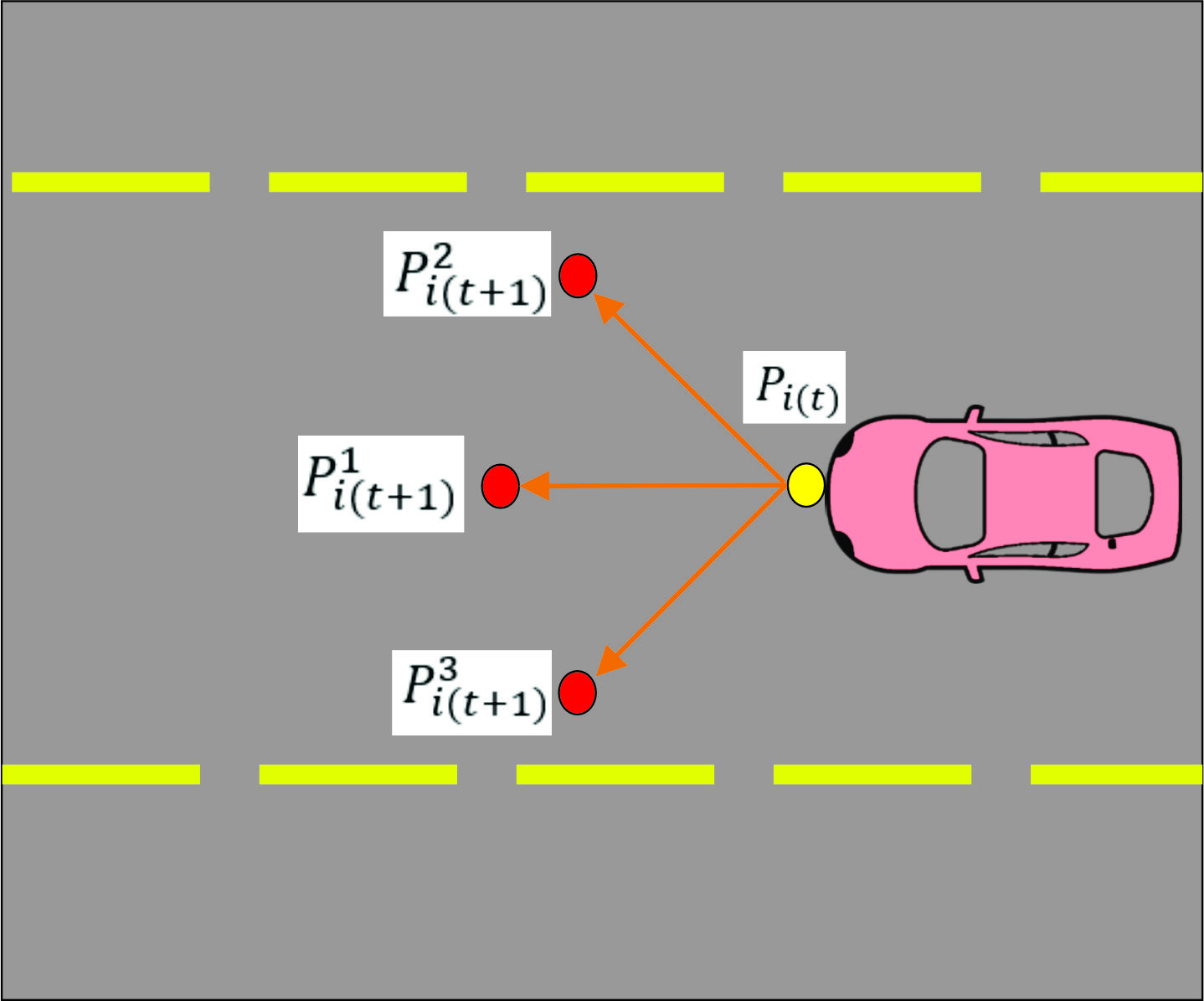}%
}

\subfloat[Multi step]{%
  \includegraphics[clip,width=0.6\columnwidth]{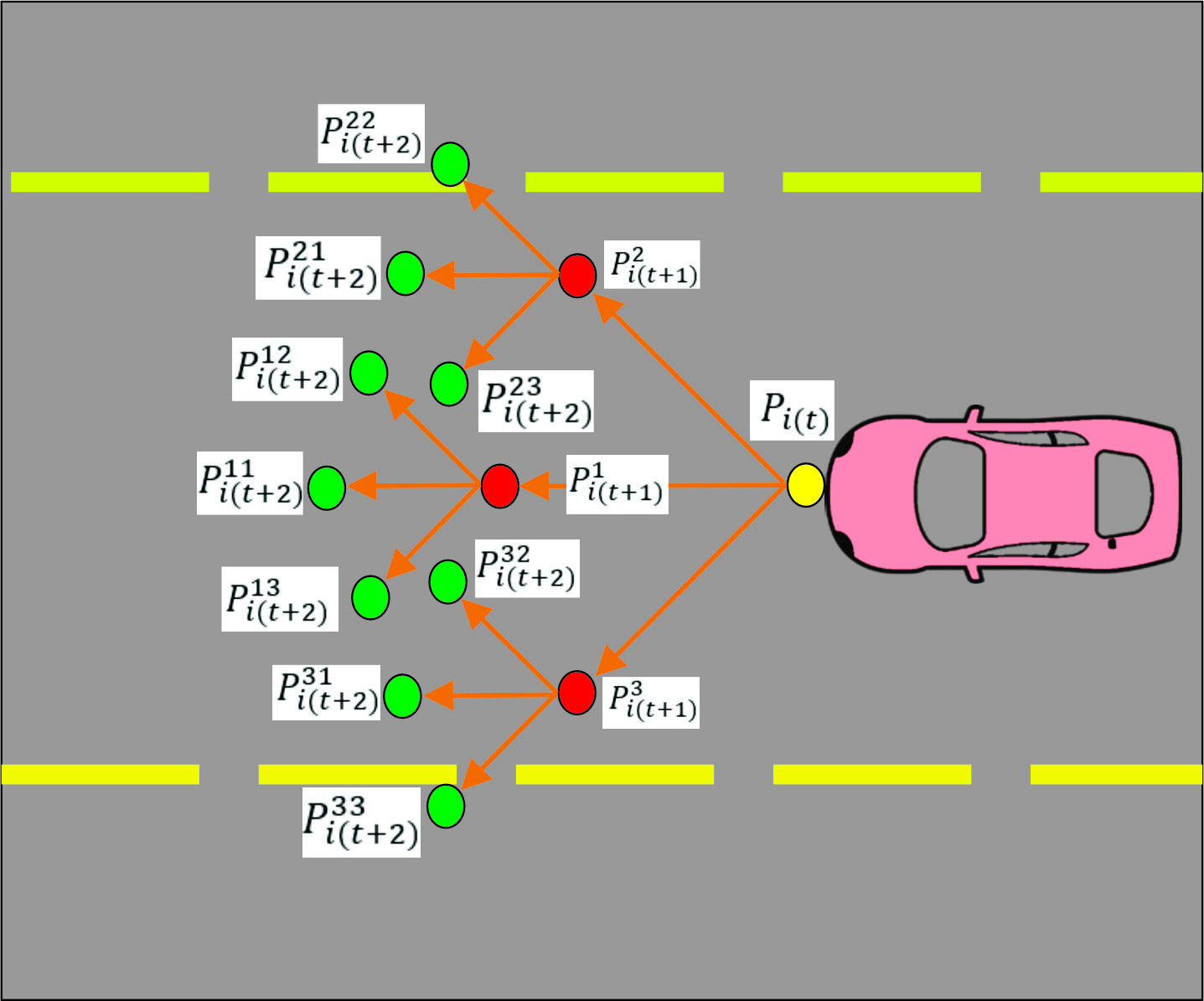}%
}
\caption{Multi and single step $N-MP$ approach}
\label{Figure: multi and single step G2MP }
\end{figure}

This section undertakes a theoretical comparison between the newly proposed $N-MP$ approach and the approach discussed in the previous section, $DMPC-MP$ approach. In the next section, the outcome of the two planners and their runtime will be investigated. Both approaches have somehow similar objectives for road users. The main difference between $DMPC-MP$ and $N-MP$ approach is that $DMPC-MP$ works in a continuous space and also considers a multi-step (MPC) approach while solving the interaction among road users. $N-MP$ approach works on a discrete space and uses the velocity of road users and some discrete changes in their orientation to find their future configurations. $N-MP$ approach relies on one step ahead of each road user to decide about interactions in the interactive traffic scenario. 
Two factors inside $N-MP$ approach make the functionality of this approach closer to $DMPC-MP$. The first factor is to consider small values for \(K_v\) and \(\delta\theta\) parameters in \eqref{length of displacement vector} and \eqref{orientation of displacement vector}, and higher value for $s_i$ in \eqref{orientation of displacement vector}. The second factor is related to the step that is investigated ahead of the current node of road user $i$, \(P_{i(t)}\). Instead of only considering one step ahead of the current node, \(P_{i(t+1)}\), and calculating the objective functions related to them, more steps can be considered and evaluated, e.g., \(P_{i(t+1)}\) and \(P_{i(t+2)}\). This modification is reflected in Figure \ref{Figure: multi and single step G2MP }. 

One of the main disadvantages of $DMPC-MP$ in comparison to $N-MP$ approach is the fact that it needs high computational time to provide the solution for an interactive traffic scenario, which makes using this approach hard in real-time applications. This problem will be investigated through simulation studies in the next section. On the other hand, $N-MP$ approach can address other types of equilibriums such as Bayesian Nash and Stackelberg equilibriums. These other types of equilibriums can be studied to tackle other considerations in real-life traffic scenarios, e.g., traffic rules, right of way, and uncertainty or loss of road users' information. Moreover, in $DMPC-MP$, the velocity and waypoints of the AV are solved in a unique optimization problem, however, in $N-MP$ approach, as the velocity and the waypoints of the AV are treated separately, there is a chance to consider different rules, logic or equilibrium for velocity and waypoints of road users.
\section{Numerical simulation}

\begin{figure}[]
  \centering
  \includegraphics[width=3cm]{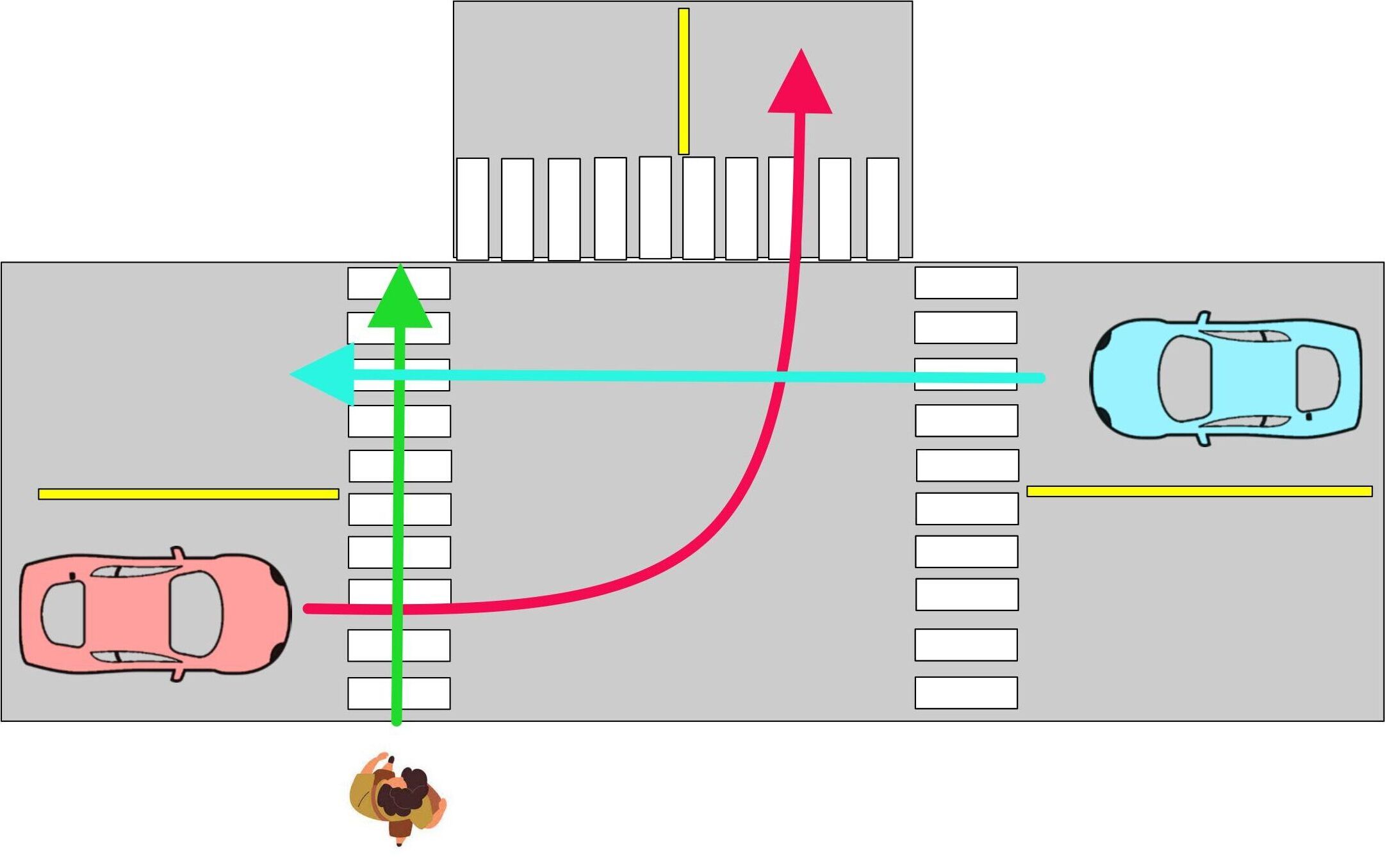}
  \caption{Simulation case.}
  \label{fig:example}
\end{figure}

In this part, an unsignalized intersection, Figure \ref{fig:example}, is considered to compare the performance of $N-MP$, $DMPC-MP$, and $ILQ-MP$ for AV navigation in complex traffic scenarios. This scenario is proposed in \cite{Fridovich2020} where a pedestrian is crossing the street while a vehicle is going straight and AV intends to make a left turn at the intersection. The model taken into account for the pedestrian's behavior assumes that he responds to both vehicles while simultaneously aiming to reach the opposite side of the road. Additionally, there is some uncertainty added to the pedestrian's movement. The human-driven vehicle, the blue vehicle, reacts laterally to both the pedestrian and AV, with a certain level of uncertainty in its motion. The AV, red vehicle, utilizes the output of three planners to safely execute the left turn. This scenario is repeated 10 times for each planner with initial positions and velocities randomly assigned to each road user for each iteration. 
Each of the three planners is individually fine-tuned to provide an optimized solution in this scenario.  The path generated by each planner is fed into the pure pursuit controller to drive the AV. The pure pursuit controller is a widely used navigation algorithm in autonomous driving systems. It operates by continuously adjusting the steering angle of the vehicle to follow a predefined path. 

\begin{figure}[htp]
  \centering
  \scriptsize
  \begin{tabular}{c}
    \includegraphics[width=2.1in]{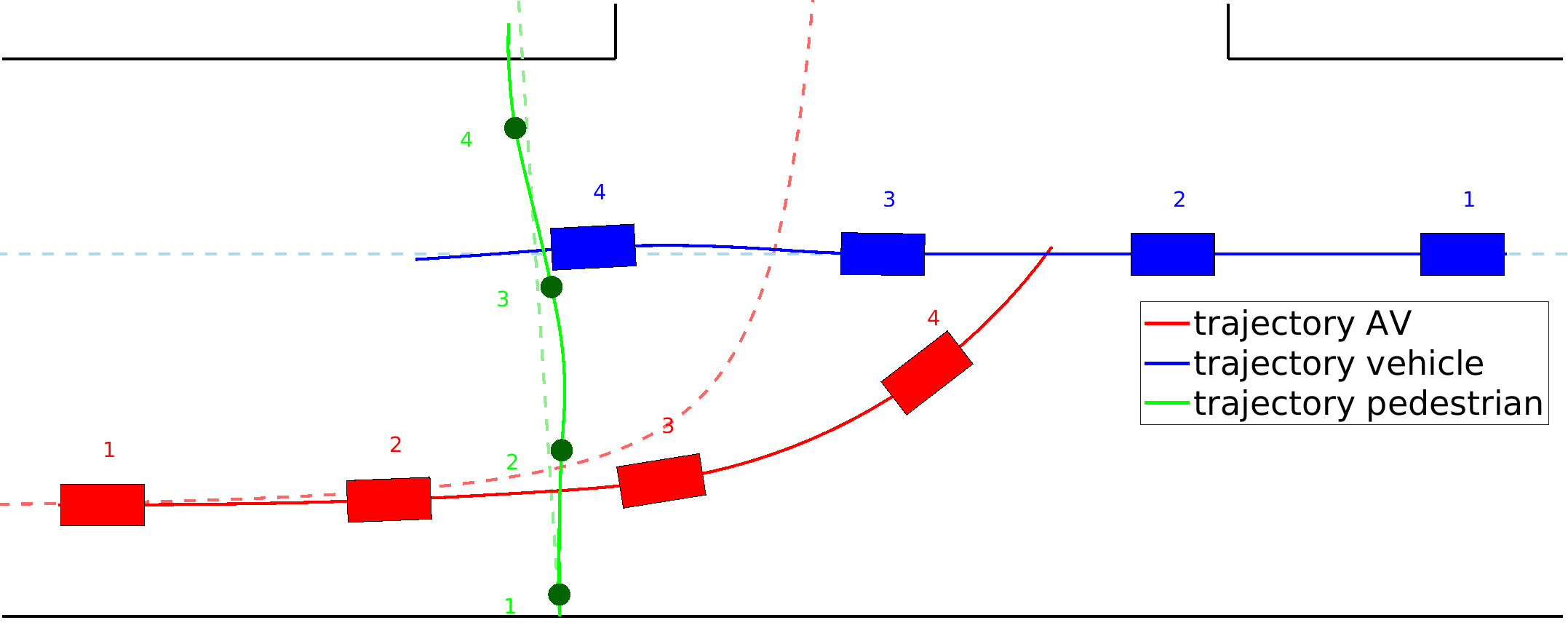}\\
    (a) complete sequence\\
    \includegraphics[width=2.1in]{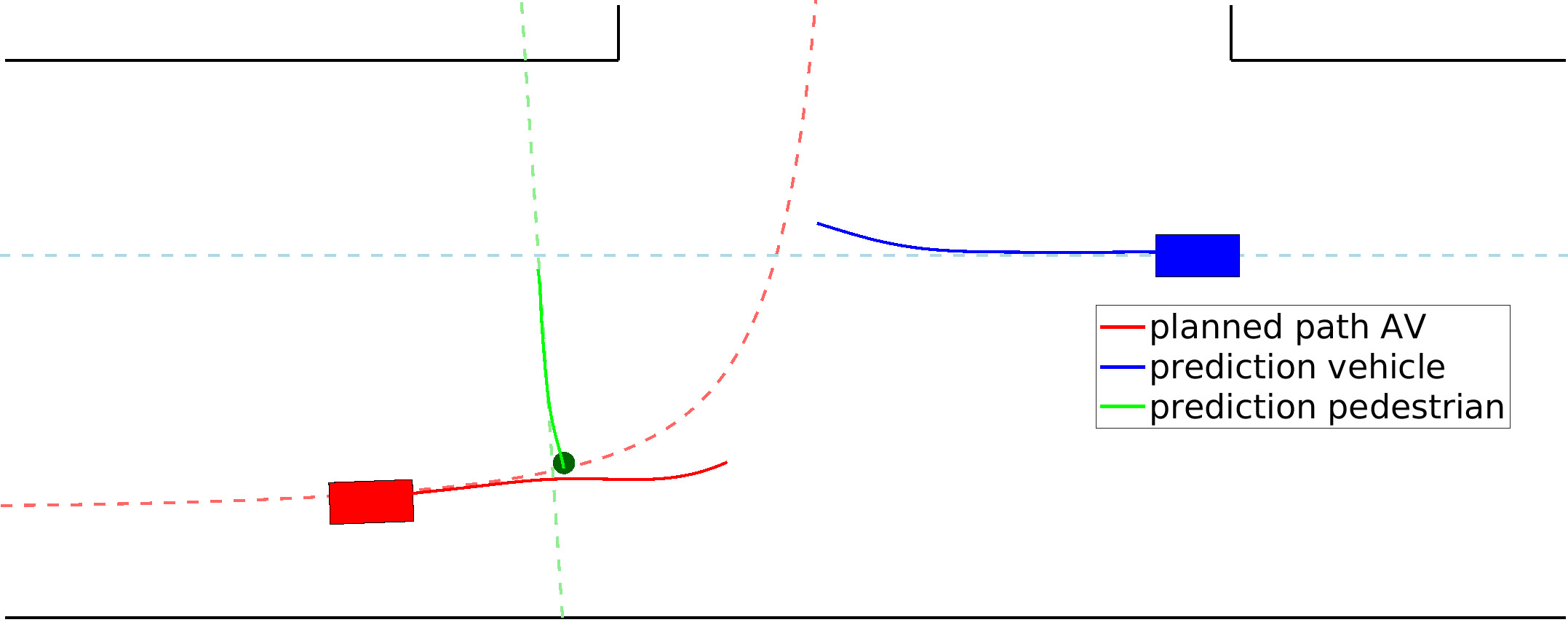}\\
    (b) planned path and prediction on $t= 1.5 (sec)$\\
    \includegraphics[width=2.1in]{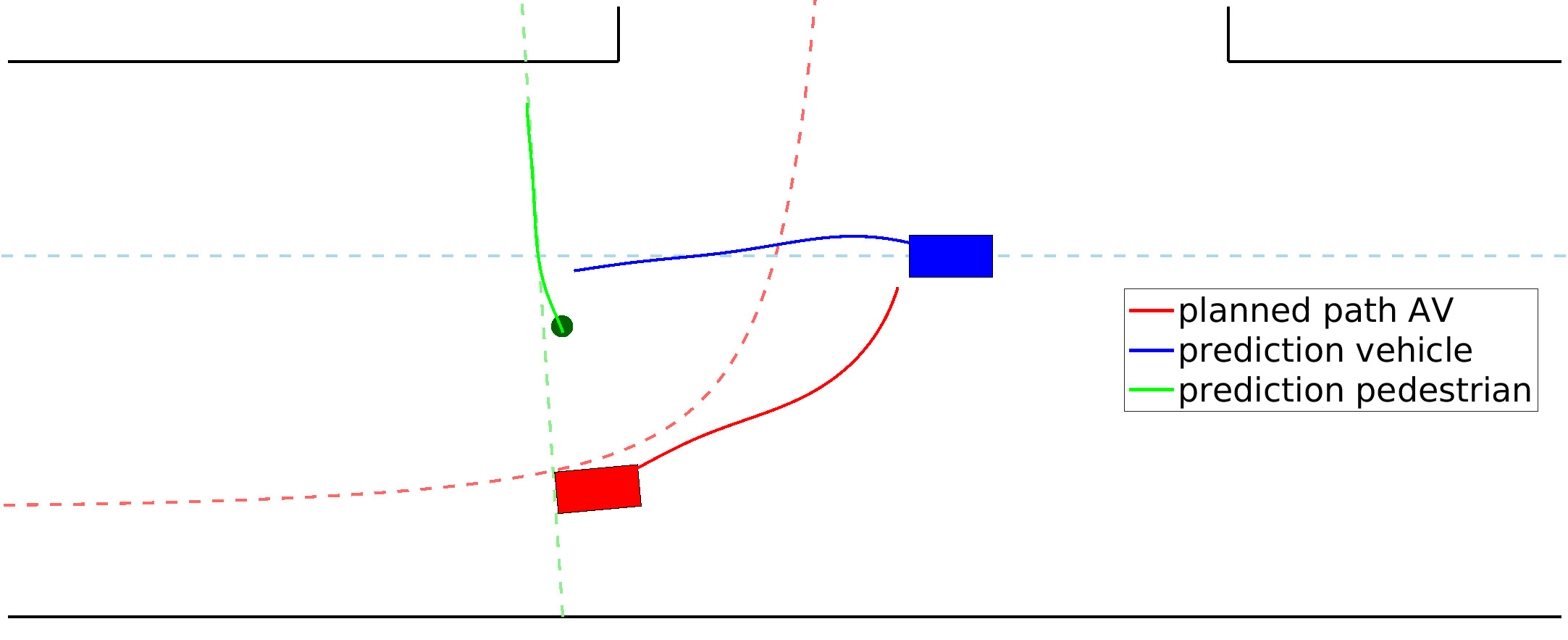}\\
    (c) planned path and prediction on $t= 2.6 (sec)$\\
  \end{tabular}
  \caption{$N-MP$}
  \label{fig:gtplanner}
\end{figure}

\begin{figure}[htp]
  \centering
\scriptsize
\begin{tabular}{c}
    \includegraphics[width=2.1in]{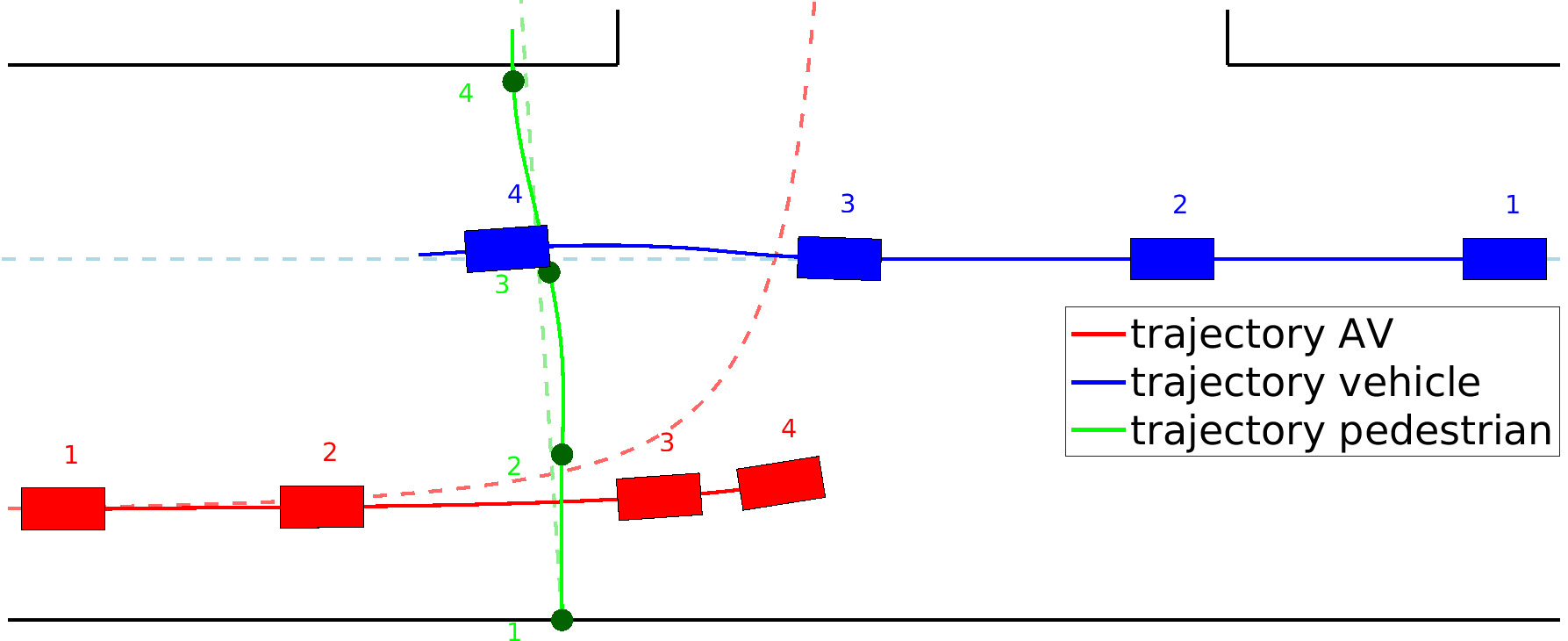}\\
    (a) complete sequence\\
    \includegraphics[width=2.1in]{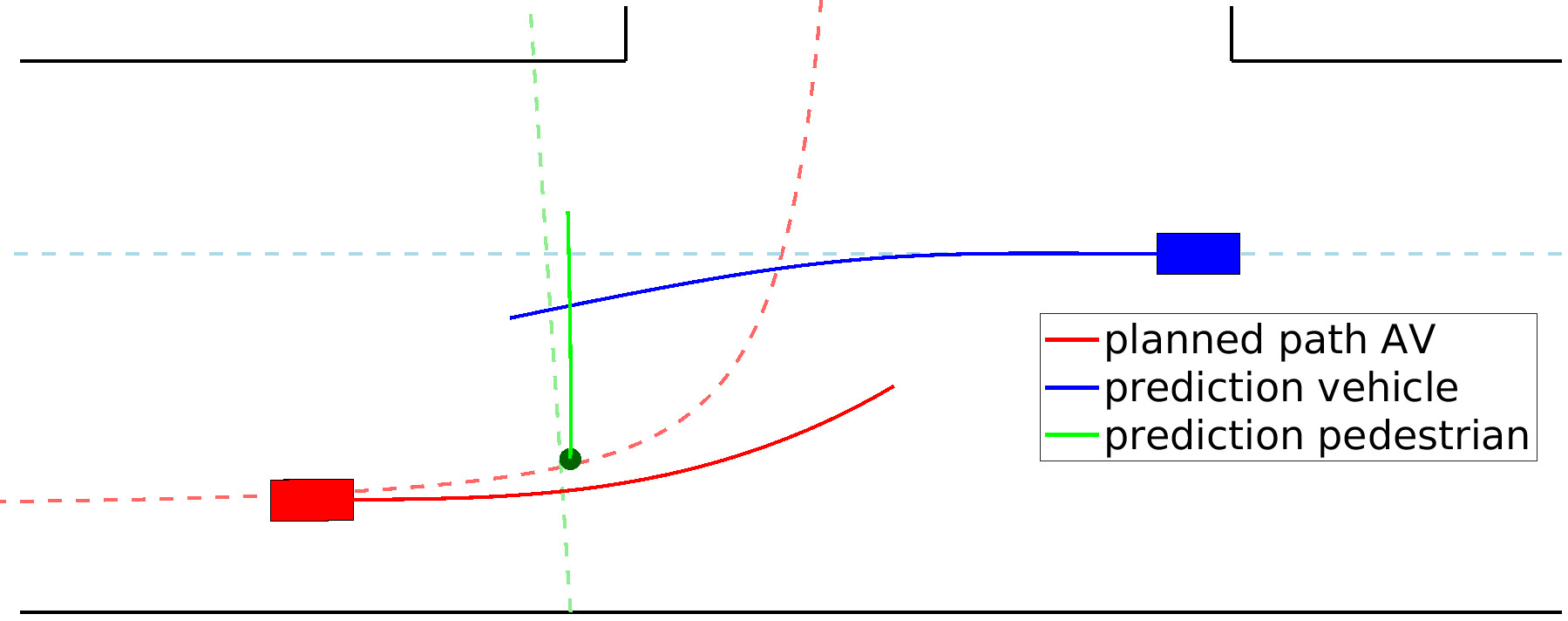}\\
    (b) planned path and prediction on $t= 1.5 (sec)$\\
    \includegraphics[width=2.1in]{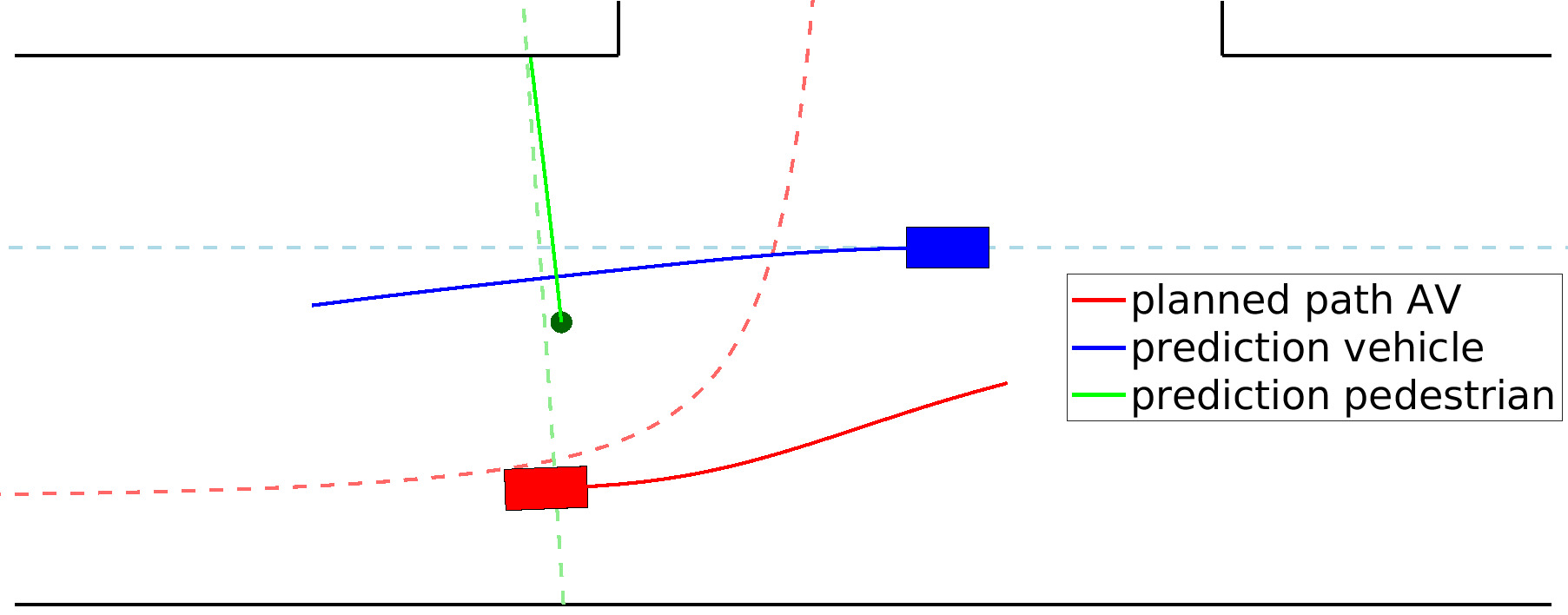}\\
    (c) planned path and prediction on $t= 2.6 (sec)$\\
  \end{tabular}
  \caption{$ILQ-MP$}
  \label{fig:ILQ}
\end{figure}

\begin{figure}[htp]
  \centering
  \scriptsize
  \begin{tabular}{c}
    \includegraphics[width=2.1in]{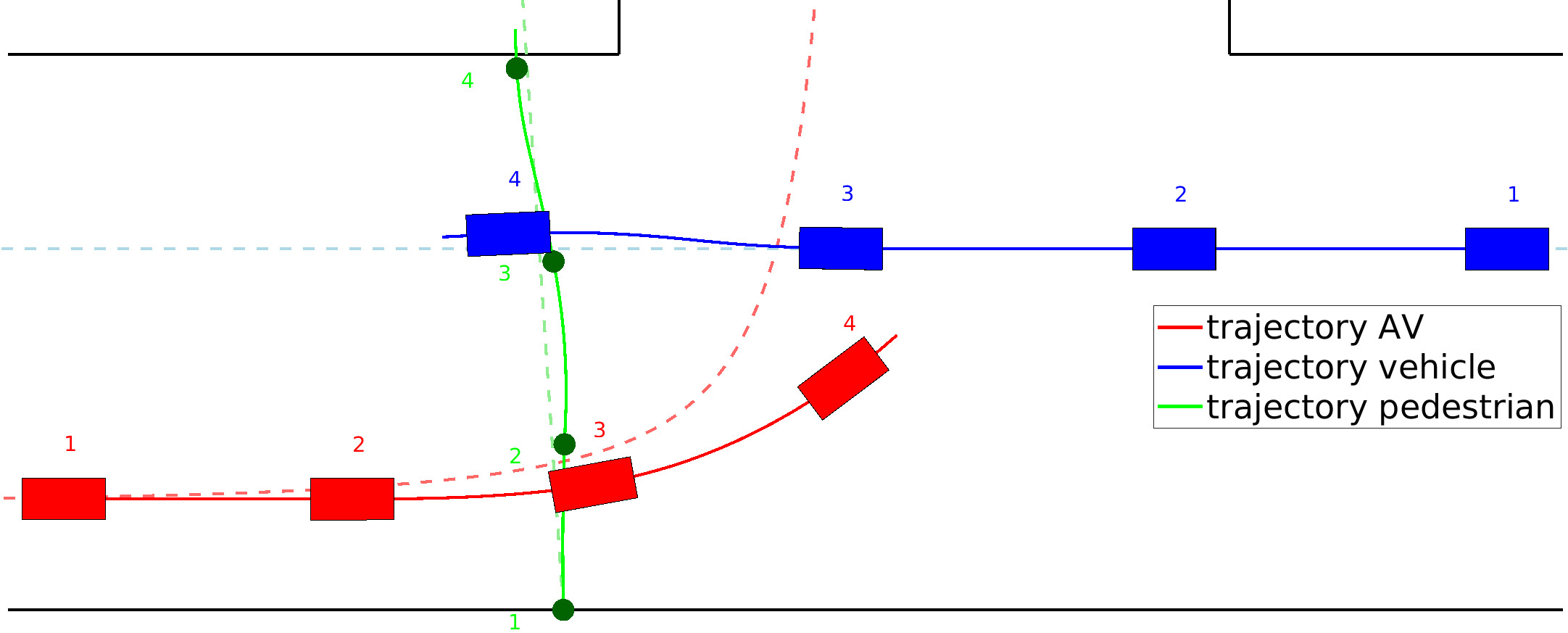}\\
    (a) complete sequence\\
    \includegraphics[width=2.1in]{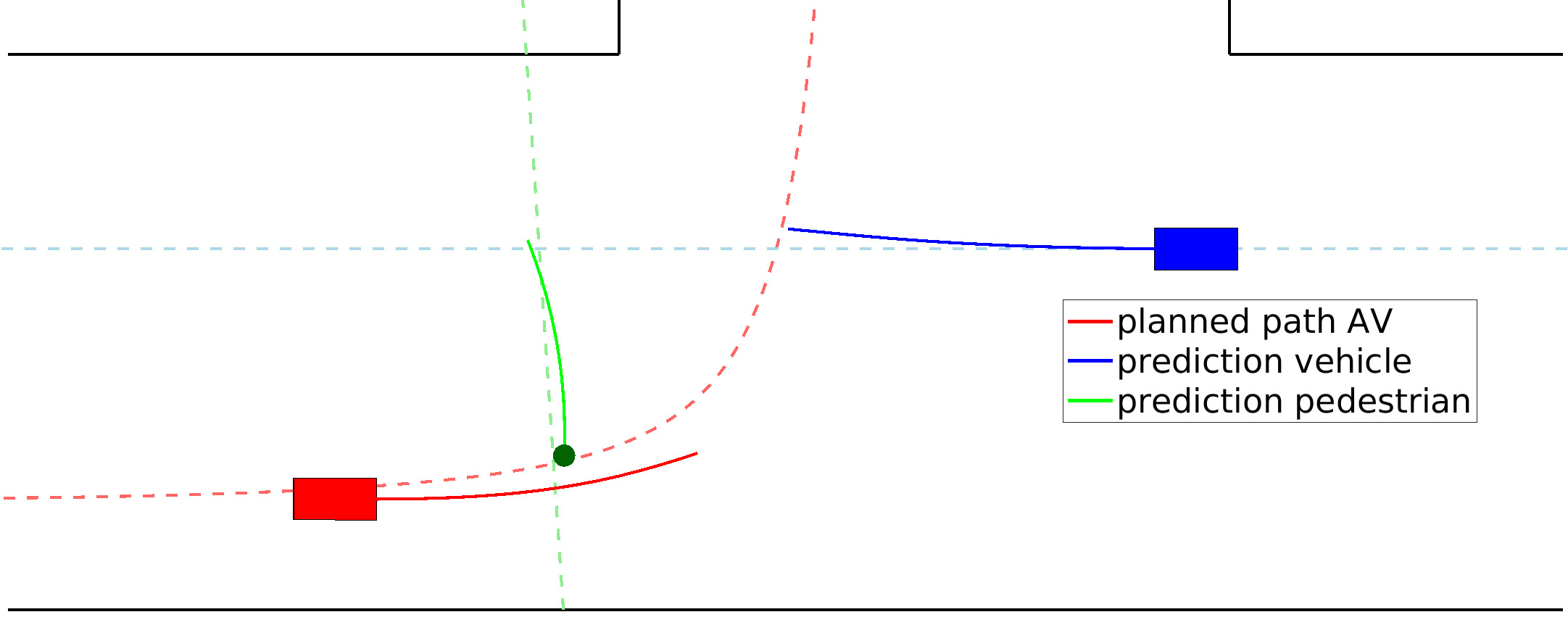}\\
    (b) planned path and prediction on $t= 1.5 (sec)$\\
    \includegraphics[width=2.1in]{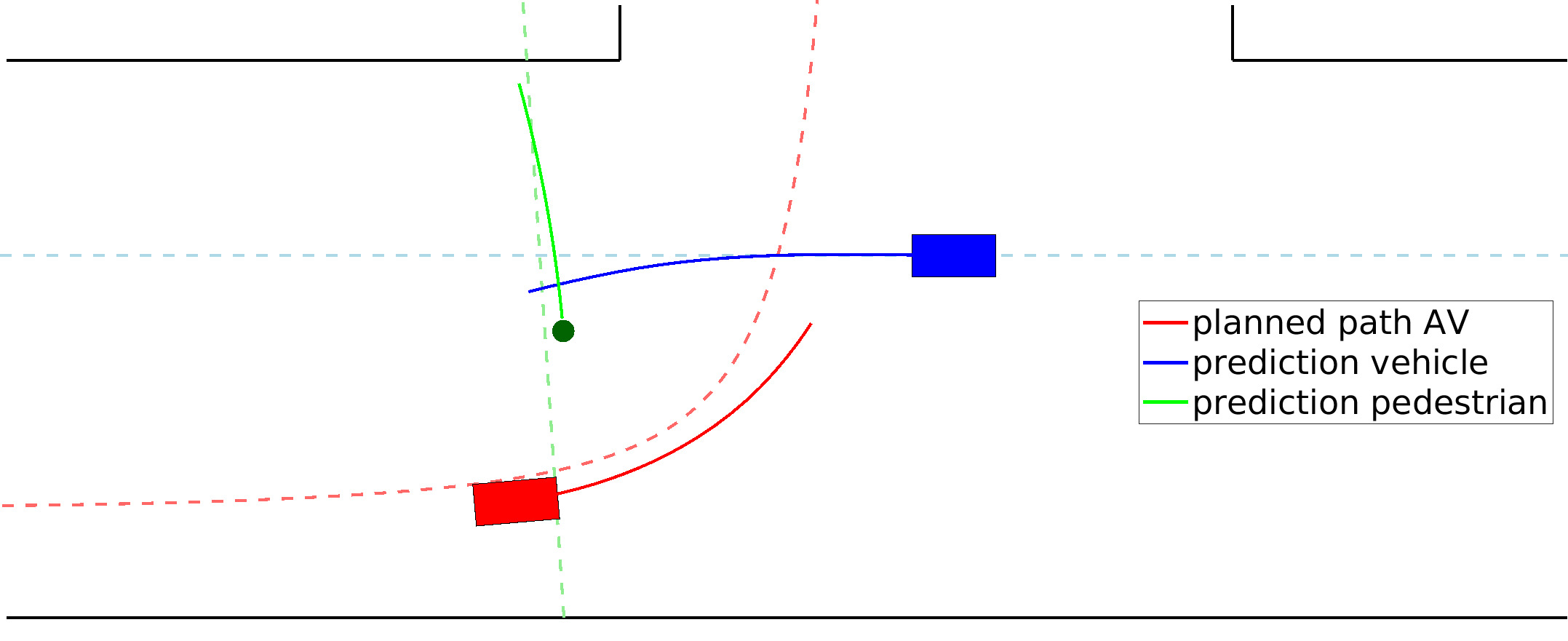}\\
    (c) planned path and prediction on $t= 2.6 (sec)$\\
  \end{tabular}
  \caption{$DMPC-MP$}
  \label{fig:DMPC}
\end{figure}

The outcomes of a single sample case, selected from a pool of 10 simulated scenarios at the intersection, are depicted in Figures \ref{fig:gtplanner}, \ref{fig:ILQ}, and \ref{fig:DMPC}, while the performance metrics of the planners are summarized in Table \ref{tab:planners}. Dash lines are reference path and solid lines are planned and predicted paths generated by planners for each road user.
The entire scenario takes approximately 5 seconds. A detailed examination of Table \ref{tab:planners} reveals that $ILQ-MP$ requires a substantial amount of time to calculate the planned path for AV. In contrast, $N-MP$ demonstrates remarkable efficiency, swiftly resolving the interaction within a shorter timeframe. Additionally, there was one case from $ILQ-MP$  where the generated path was untrackable for the AV.
\begin{table}[]
    \centering
    \caption{Performance of planners in intersection scenario}
    \label{tab:planners}
    \scriptsize
    \begin{tabular}{>{\raggedright}p{4cm}p{1cm}p{1cm}p{1cm}} 
        \toprule
        \textbf{Planner} & \textbf{ILQ} & \textbf{Nash} &  \textbf{DMPC} \\
        \midrule
        Average time (sec) & 178& 1.01 & 46.45\\
        \addlinespace
        Divergence (no optimal solution) & 1 & 0 & 0\\
        \bottomrule
    \end{tabular}
\end{table}

Considering the planned and predicted paths from the three planners at $t=1.5$ seconds, part (b) in Figures \ref{fig:gtplanner}, \ref{fig:ILQ}, and \ref{fig:DMPC}, it becomes evident that all of them veer the path of AV, red path, towards the right side of the AV to evade collision with the pedestrian. Interestingly, both $N-MP$ and $DMPC-MP$ anticipate that the human-driven vehicle will also shift its path, blue path, towards its right side to evade the pedestrian. However, $ILQ-MP$ predicts that the human-driven vehicle will deviate from its reference path toward the left to avoid the pedestrian.
At $t=2.6$ seconds, part (c) in Figures \ref{fig:gtplanner}, \ref{fig:ILQ}, and \ref{fig:DMPC}, as the AV passes the pedestrian and approaches the vicinity of the human-driven vehicle, it becomes apparent that the planned path from three planners, red path, instruct the AV to position itself to the right of its reference path to avoid human-driven vehicle. However, it is notable that the path generated by $DMPC-MP$ closely aligns with the AV's reference path, whereas the path produced by $ILQ-MP$ deviates the most from it. Furthermore, each planner predicts that the human-driven vehicle will maneuver towards its left side to avoid the pedestrian.
These simulations offer insights indicating that the planning and predictive capabilities of $N-MP$ closely resemble those of $DMPC-MP$ and $ILQ-MP$. However, a notable distinction lies in the efficiency with which $N-MP$ generates outcomes and completes the task in a significantly shorter duration. This observation proves the agility and effectiveness of $N-MP$ in swiftly navigating complex scenarios, contributing to its potential as a promising solution for real-time autonomous driving applications.

\begin{figure}[H]
    \centering
    \includegraphics[width=0.35\linewidth]{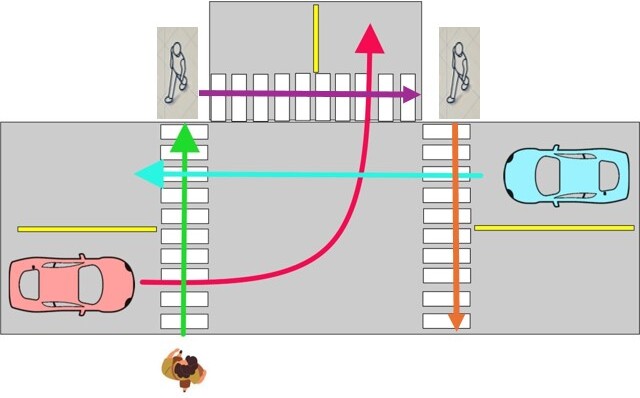} 
    \caption{Simulation scenario with 5 road users}
    \label{fig:ComplexSimulation}
\end{figure}

\begin{figure}[H]
    \centering
    \includegraphics[width=0.7\linewidth]{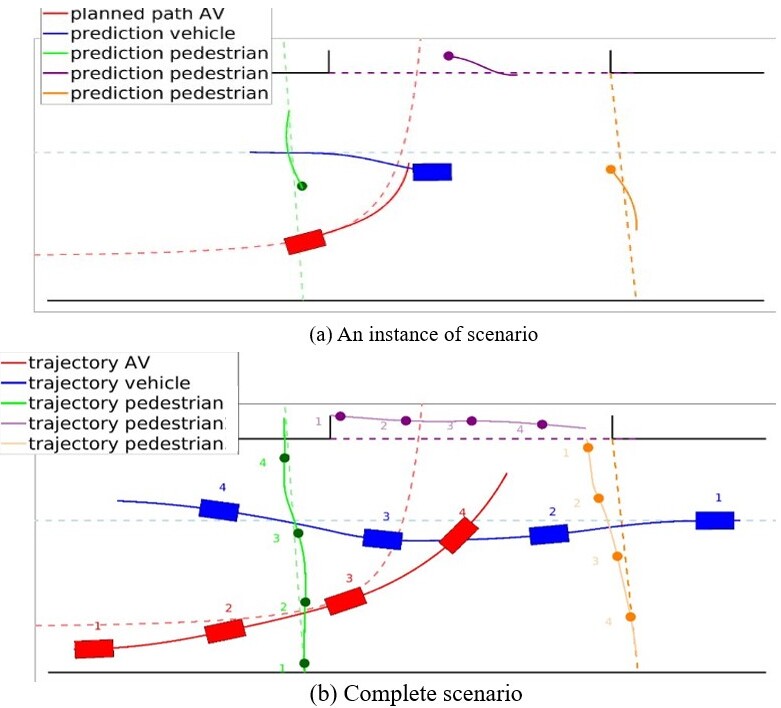} 
    \caption{Simulation result for the scenario with 5 road users}
    \label{fig:ComplexScenarioResult}
\end{figure}

Finally, to evaluate the performance of $N-MP$ in scenarios with a higher number of road users, we introduce two additional pedestrians into the previous setup, as shown in Figure \ref{fig:ComplexSimulation}. As the number of road users increases, the computational time for solving the entire interaction and planning the AV's motion rises to 3 seconds. Given that the entire scenario takes approximately 5 seconds to play out, the planner's performance remains within the real-time domain, allowing it to handle the increased complexity effectively. Both the full scenario and a representative instance are illustrated in Figure \ref{fig:ComplexScenarioResult}.

\section{Experimental testing}

\begin{figure}[H]
  \centering
  \begin{tabular}{cc}
    \includegraphics[width=1.5in]{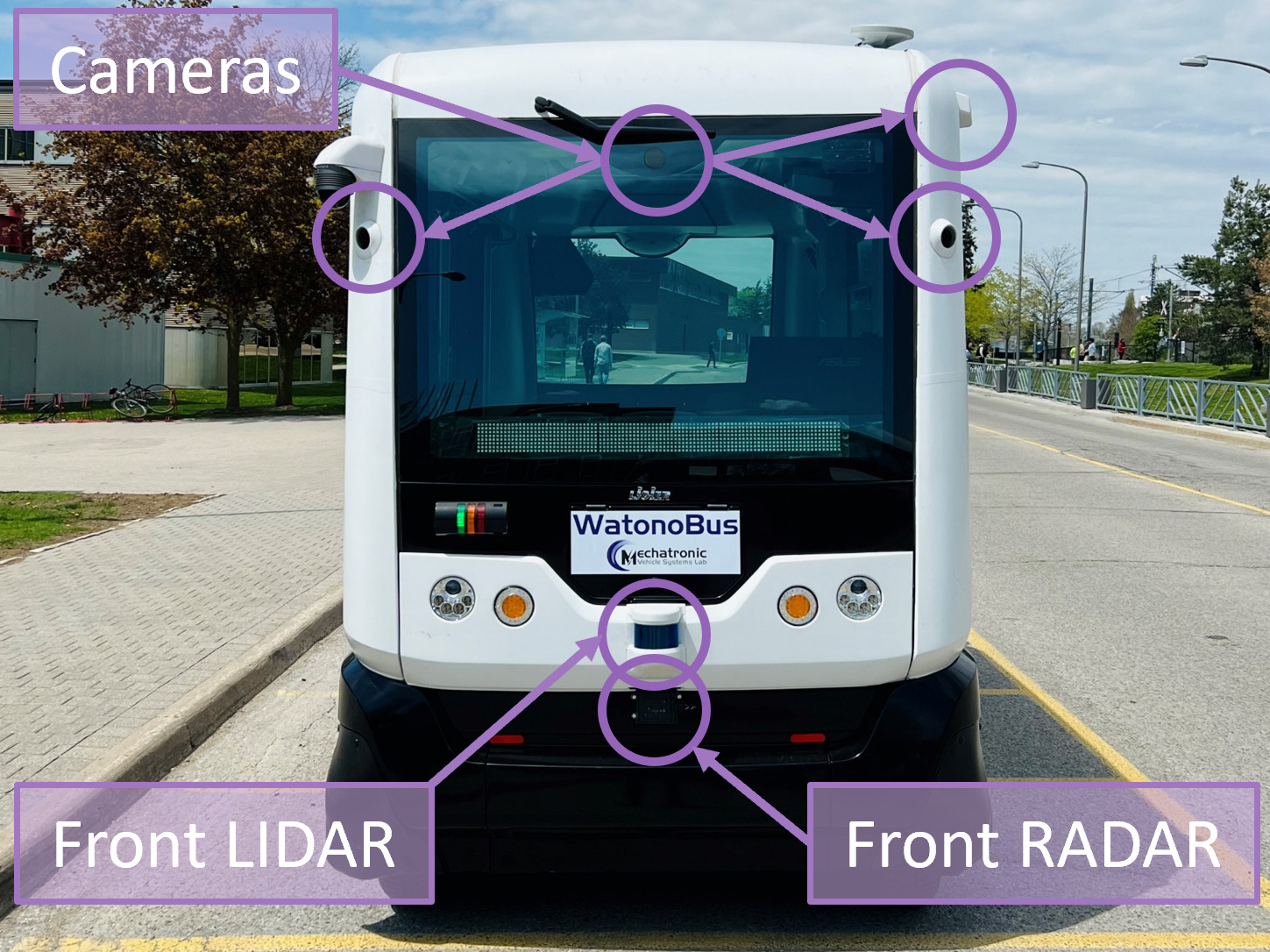} &
    \includegraphics[width=1.5in]{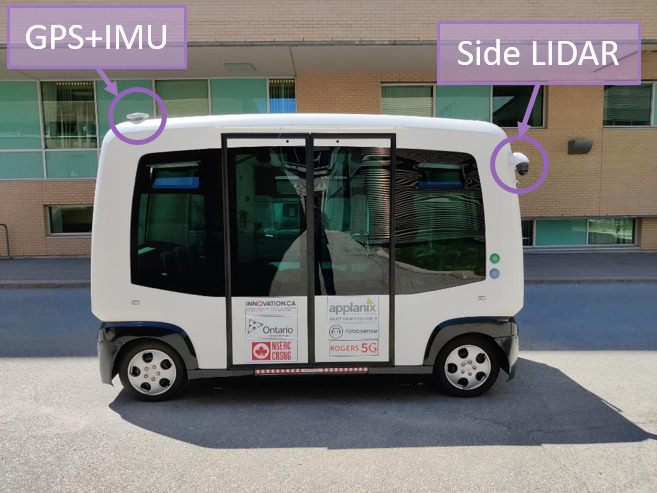}\\
  \end{tabular}
  \caption{WATonoBus, driven by Mechatronics Vehicles System Laboratory at the University of Waterloo. The vehicle benefits from double steering and a maximum speed of 20 kph.}
  \label{ShuttleBus}
\end{figure}

To validate the effectiveness of $N-MP$ in real-world scenarios, we conducted an experiment focusing on a situation where WATonoBus (depicted in Figure \ref{ShuttleBus}) encounters a merging vehicle near a bus stop. During this experiment, conducted on a rainy day, as the WATonoBus approached its designated stop, a parked vehicle unexpectedly attempted to merge onto the main road. This sudden maneuver presented a complex challenge for the WATonoBus to navigate safely.
The WATonoBus comes equipped with a comprehensive array of sensory equipment to provide a thorough understanding of its surroundings. It is equipped with three forward-facing cameras and a rear camera, providing complete visual coverage. Additionally, it features two blind spot LiDAR sensors, enhancing awareness of nearby obstacles. Augmenting this perception, the bus integrates a front LiDAR unit and a radar system, ensuring robust environmental sensing capabilities. This advanced perception system delivers detailed information about the position, orientation, and velocity of other road users.
To accurately determine its position, orientation and velocity along the University ringroad, the WATonoBus relies on an Applanix GPS system, ensuring precise localization throughout its navigation. The path generated from $N-MP$ is inputted into a MPC controller to drive the movement of the WATonoBus.

\begin{figure}[H]
    \centering
    \includegraphics[width=0.6\linewidth]{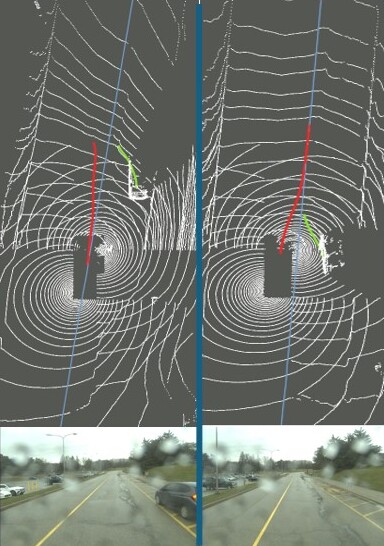} 
    \caption{Experiment scene: Rviz(up) and camera view(bottom). Red line: planned path for WATonoBus. Green line: predicted path for vehicle. Blue line: reference path.}
    \label{fig:experimentRvizCamera}
\end{figure}

\begin{figure}[h]
    \centering
    \includegraphics[width=0.8\linewidth]{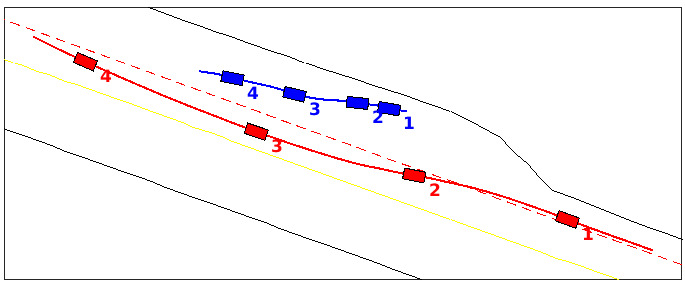} 
    \caption{Experiment scenario. WATonoBus: red. Human-driven vehicle: blue.}
    \label{fig:experiment}
\end{figure}

In this experiment, we aim to test the real-time planning and prediction capabilities of $N-MP$ for both WATonoBus and human-driven vehicle. The real-time performance of  $N-MP$ is illustrated in Figure \ref{fig:experimentRvizCamera}, which provides views from Rviz and the front camera captured at two different moments. The red solid line is the path planned for WATonoBus and solid green line is the predicted path for human-driven vehicle.
Initially, $N-MP$ guided the WATonoBus to the left side, ensuring a safe distance from the merging human-driven vehicle. Subsequently, as the WATonoBus passed the human-driven vehicle, $N-MP$ adjusted the vehicle's trajectory back towards the reference path, solid blue line.
The complete sequence of events is depicted in Figure \ref{fig:experiment}. This experiment highlights the capability of $N-MP$ to operate in real-time and effectively guide WATonoBus through challenging interactions with other road users.
Testing the motion planner presents a challenge due to potential inaccuracies in determining the center position of the human-driven vehicle, which can impact the planner's performance. In Figure \ref{fig:experimentRvizCamera}, the left scene illustrates the perception system locating the center of the human-driven vehicle somewhere near the rear of the vehicle, while the right scene depicts it reporting the center somewhere near the front left of the vehicle. This challenge will be an essential consideration for future algorithm implementations.

\section{Conclusion}
This study introduces a novel approach to investigating the interaction between AV and road users. The approach utilizes game theory and search-based  techniques to formulate and solve the interactive motion planning problem formed between AV and road users. The movement of each road user in the traffic scenario is integrated into the game using graph-defined state space transitions, with the length of the graphs reflecting the speed of the road users. The approach considers the reactions of other road users when determining the future trajectory of the AV, and the trajectories are derived based on Nash equilibria involving all road users in the traffic scenario. In comparison to the other interactive motion planners, the developed framework has lower computational complexity, making it suitable for real-time applications. 

Our work can be extended in several ways. 
In future studies, our aim is to explore alternative equilibriums, such as Stackelberg and Bayesian equilibrium, to effectively address the challenges posed by traffic regulations and uncertainties encountered in everyday traffic situations. By investigating these equilibrium frameworks, we seek to enhance the AV's motion planning strategies while considering the strategic interactions with other road users and incorporating probabilistic models for handling uncertainties. These investigations hold promise for advancing our understanding of traffic dynamics and enabling the development of more sophisticated and robust approaches to AV motion planning in real-world scenarios.
\section*{Acknowledgment}
The authors would like to acknowledge the financial support of the Natural Sciences and Engineering Research Council of Canada (NSERC) in this work. 
\ifCLASSOPTIONcaptionsoff
  \newpage
\fi

\scriptsize
\bibliography{Ref}
\bibliographystyle{unsrt}

\scriptsize
\begin{IEEEbiography}[{\includegraphics[width=1in,height=1.35in,clip,keepaspectratio]{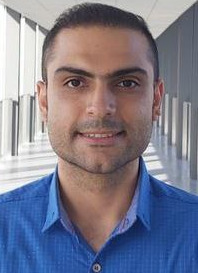}}]{Pouya Panahandeh}
is currently pursuing his Ph.D. degree in the Department of Mechanical and Mechatronics Engineering at the University of Waterloo. His primary area of research interest is motion planning, prediction, and the development of safe autonomous driving systems. Specifically, he is exploring techniques to enhance the efficiency of real-time motion planning algorithms in diverse urban traffic situations. Pouya's investigations are part of the ongoing WATonoBus autonomous shuttle project, which aims to improve the accuracy and reliability of autonomous driving in dynamic environments.
\end{IEEEbiography}
\begin{IEEEbiography}[{\includegraphics[width=1in,height=1.25in,clip,keepaspectratio]{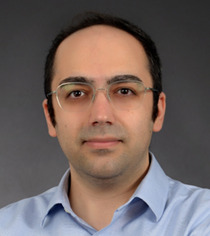}}]{Mohammad Pirani}
is a research assistant professor in the Department of Mechanical and Mechatronics  Engineering at the University of Waterloo. He held postdoctoral researcher positions at the University of Toronto (2019-2021) and  KTH Royal Institute of Technology, Sweden (2018-2019). He received a MASc degree in electrical and computer engineering and a Ph.D. degree in Mechanical and Mechatronics Engineering, both from the University of Waterloo in 2014 and 2017, respectively. His research interests include resilient and fault-tolerant control, networked control systems, and multi-agent systems.
\end{IEEEbiography}
\begin{IEEEbiography}[{\includegraphics[width=1in,height=1.35in,clip,keepaspectratio]{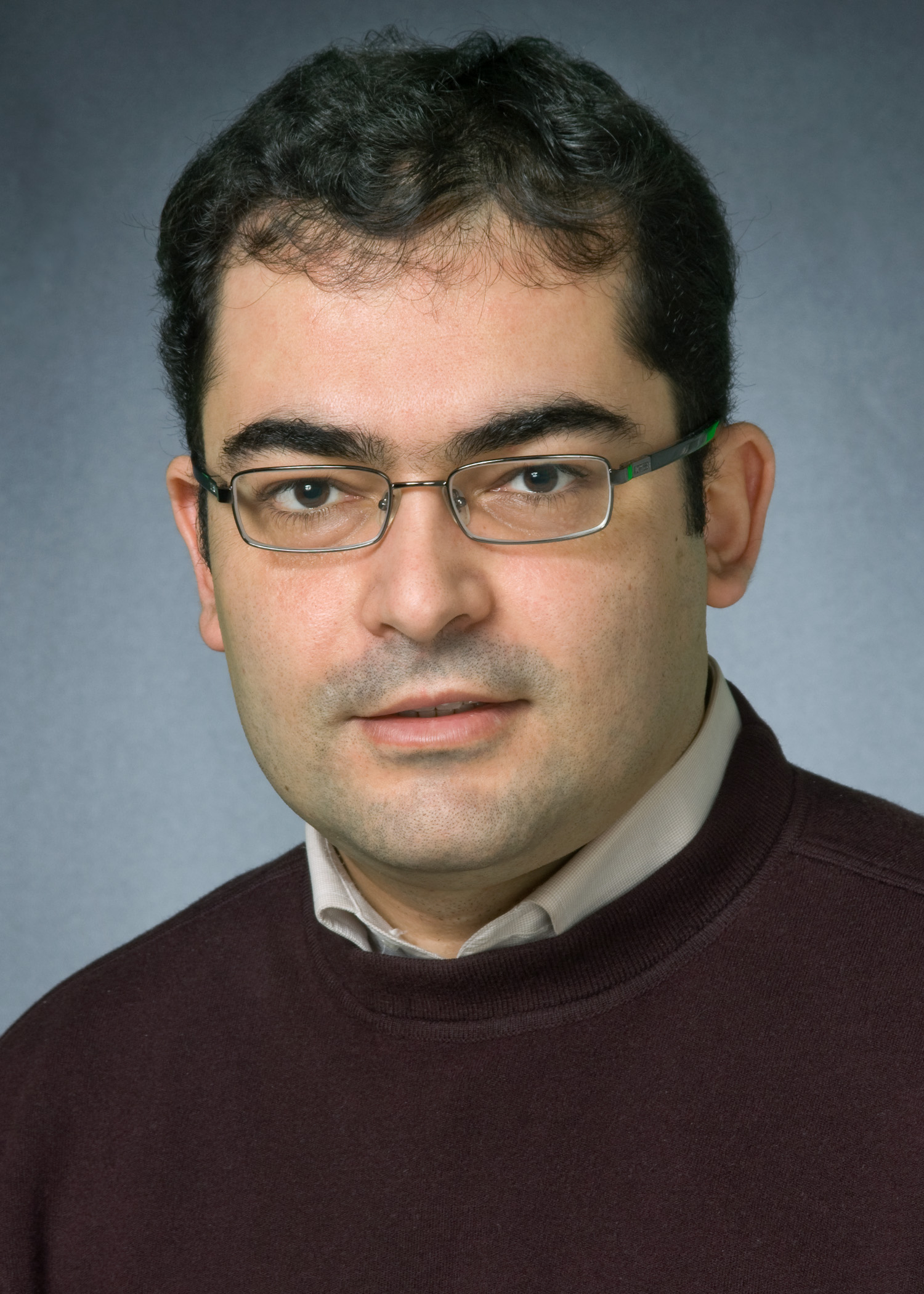}}]{Baris Fidan}
received the B.S. degrees in electrical engineering and mathematics from  Middle East Technical University, Turkey, the M.S. degree in electrical engineering from Bilkent University, Turkey, and the Ph.D. degree in electrical engineering at the University of Southern California, USA in 2003. He worked at the University of Southern California in 2004 as a postdoctoral research fellow, and at the National ICT Australia and the Research School of Information Sciences and Engineering of the Australian National University in 2005-2009 as researcher/senior researcher. He is currently a professor at the Mechanical and Mechatronics Engineering Department, University of Waterloo, Canada. His research interests include system identification and adaptive control, switching and hybrid systems, multi-vehicle system coordination, autonomous and intelligent vehicle systems, sensor networks, cooperative and adaptive extremum seeking, human assistive robotics, and various control applications including vehicle stability control, high performance and UAV flight control, intelligent manufacturing, and industrial robotics.
\end{IEEEbiography}
\begin{IEEEbiography}[{\includegraphics[width=1in,height=1.3in,clip,keepaspectratio]{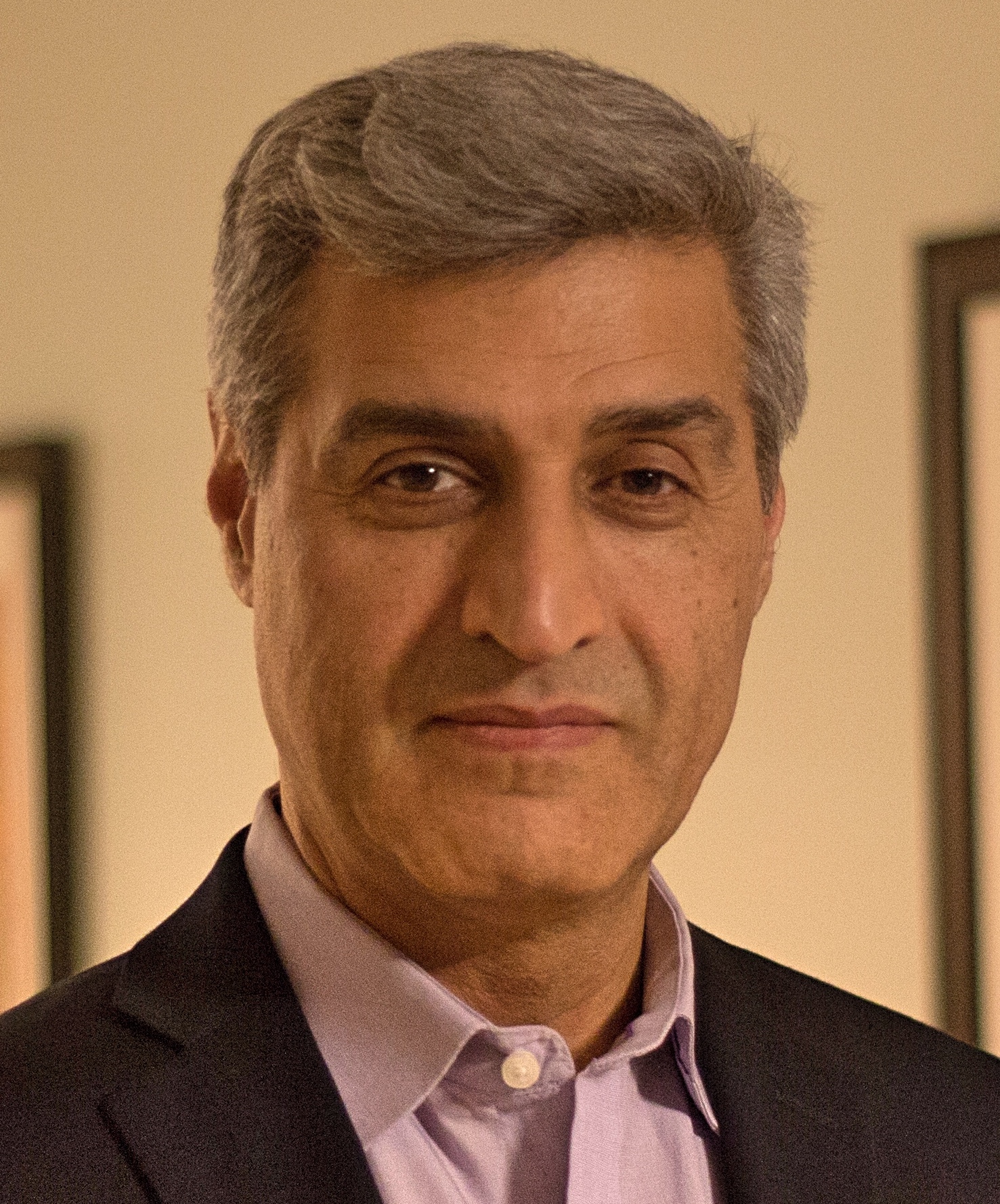}}]{Amir Khajepour}
(Senior Member, IEEE) held the
Tier 1 Canada Research Chair in Mechatronic Vehicle Systems from 2008 to 2022 and the Senior NSERC/General Motors Industrial Research Chair in Holistic Vehicle Control from 2017 to 2022. He is currently a Professor of mechanical and mechatronics engineering and the Director of the Mechatronic Vehicle Systems (MVS) Laboratory, University of Waterloo. His work has resulted in training of over 150 Ph.D. and M.A.Sc. students, 30 patents, more than 600 research articles, numerous technology transfers, and several start-up companies. He is a fellow of the Engineering Institute of Canada, the American Society of Mechanical Engineering, and the Canadian Society of Mechanical Engineering. He has been recognized with the Engineering Medal from Professional Engineering Ontario.
\end{IEEEbiography}

\end{document}